\documentclass[letterpaper, 10 pt, conference]{ieeeconf}  

\IEEEoverridecommandlockouts                              
\usepackage[utf8]{inputenc}
\usepackage[T1]{fontenc}
\usepackage{amsfonts}

\usepackage[table]{xcolor}
\usepackage{booktabs}
\usepackage{multirow}
\usepackage{makecell}
\usepackage{xcolor}
\usepackage{pifont}
\usepackage{tabularx}
\usepackage{array}
\usepackage{arydshln}
\usepackage{dashrule}
\usepackage{pifont}

\definecolor{rblue}{rgb}{0,0.5,1}
\definecolor{awesome}{rgb}{1.0, 0.13, 0.32}
\definecolor{hollywoodcerise}{rgb}{0.96, 0.0, 0.63}
\definecolor{lasallegreen}{rgb}{0.03, 0.47, 0.19}
\definecolor{hanpurple}{rgb}{0.32, 0.09, 0.98}
\definecolor{green(pigment)}{rgb}{0.0, 0.65, 0.31}
\definecolor{ceilingcolor}{HTML}{E62F35}
\definecolor{floorcolor}{HTML}{25A534}
\definecolor{wallcolor}{HTML}{9BD6E2}
\definecolor{windowcolor}{HTML}{7399CF}
\definecolor{chaircolor}{HTML}{C4C943}
\definecolor{bedcolor}{HTML}{FFB168}
\definecolor{sofacolor}{HTML}{8F6AB0}
\definecolor{tablecolor}{HTML}{2074B8}
\newcommand{\cmark}{\textcolor{red!85!black}{\ding{51}}}
\newcommand{\xmark}{\textcolor{green!60!black}{\ding{55}}}
\newcommand{\xmarkblack}{\textcolor{black!60!black}{\ding{55}}}
\newcommand{\cmarkblack}{\textcolor{black!85!black}{\ding{51}}}
\DeclareRobustCommand{\perspicon}{%
    \raisebox{-0.2\height}{%
        \includegraphics[height=2.0ex]{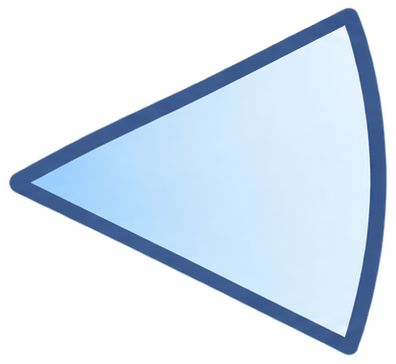}%
    }%
}
\DeclareRobustCommand{\panoicon}{%
    \raisebox{-0.2\height}{%
        \includegraphics[height=2.0ex]{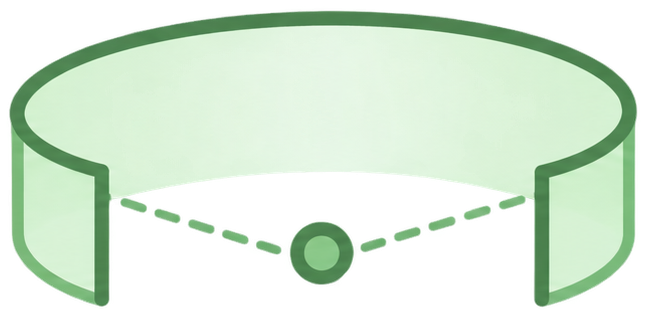}%
    }%
}

\makeatletter
\let\NAT@parse\undefined
\makeatother
\usepackage[pagebackref=false, breaklinks=true, colorlinks, bookmarks=false]{
        hyperref
}
\hypersetup{
        colorlinks=true,
        linkcolor={red},
        citecolor={hanpurple},
        urlcolor={magenta}
}

\usepackage{capt-of}
\usepackage{graphicx}
\usepackage{amsmath,amssymb}

\title{\LARGE \bf
PanOVOcc: Panoramic Embodied Open-Vocabulary Occupancy Mapping with Long-term Spatial Voxel Memory
}

\author{Di Kuang$^{1}$, Mengfei Duan$^{1}$, Yuhang Wang$^{1}$, Weixing Peng$^{1,2,\dag}$, and Kailun Yang$^{1,2,\dag}$
\thanks{This work was supported in part by National Natural Science Foundation of China under Grant No. 62573181, No. 62293512, No. 62473139, and No. 62388101, in part by the Hunan Natural Science Foundation under Grant 2026JJ40061, in part by the Hunan Provincial Research and Development Project (Grant No. 2025QK3019), and in part by the State Key Laboratory of Autonomous Intelligent Unmanned Systems (the opening project number ZZKF2025-2-10).}
\thanks{$^{1}$The authors are with the School of Artificial Intelligence and Robotics, Hunan University, China (email: kailun.yang@hnu.edu.cn).}%
\thanks{$^{2}$The authors are also with the National Engineering Research Center of Robot Visual Perception and Control Technology, Hunan University, China.}%
\thanks{$^{\dag}$Corresponding authors: Weixing Peng and Kailun Yang.}
}

\let\oldtwocolumn\twocolumn
\renewcommand\twocolumn[1][]{%
    \oldtwocolumn[{#1}{
    \begin{center}
    \vskip -3ex
        \centering
        \includegraphics[width=0.95\textwidth]{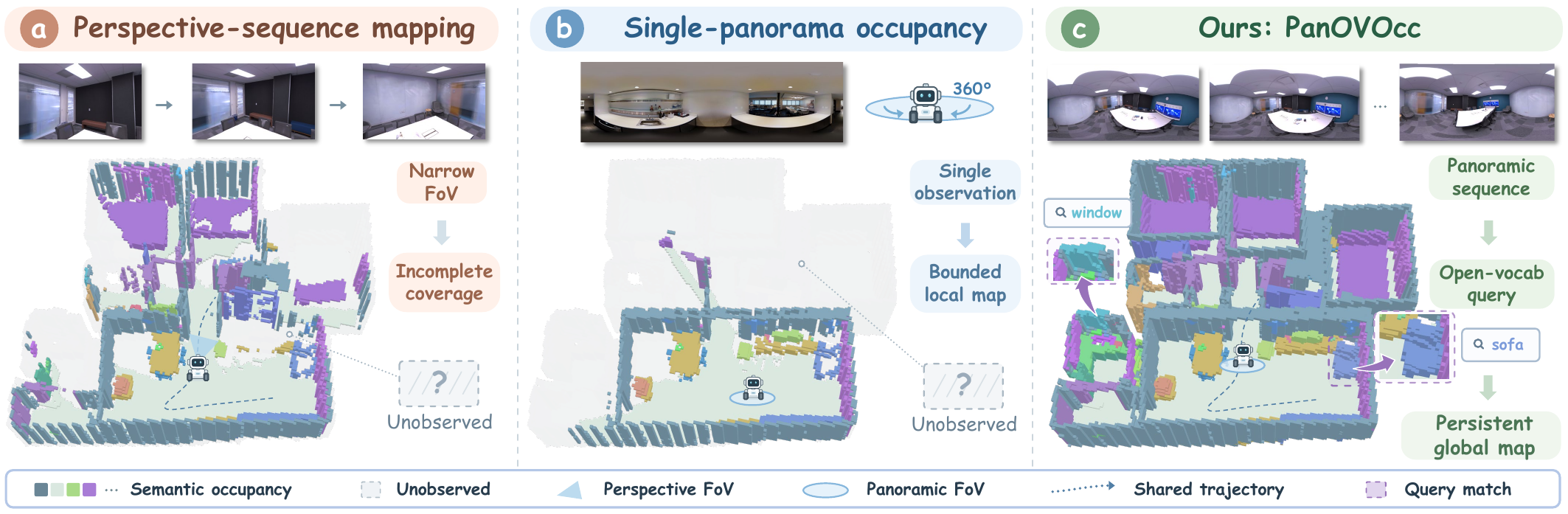}
        \vskip -1ex
        \captionof{figure}{{\textbf{Comparison of three mapping paradigms under a unified scene representation.} 
        (a) Perspective-sequence mapping integrates temporally ordered observations into a persistent global map, but suffers from incomplete spatial coverage. 
        (b) Single-panorama occupancy mapping provides omnidirectional coverage, yet only produces a local map without persistent global memory. 
        (c) In contrast, our proposed PanOVOcc continuously integrates a sequence of panoramic images into a globally consistent spatial memory, enabling both complete omnidirectional mapping and open-vocabulary semantic querying.
        }}
        \label{fig:teaser}
    \end{center}
    }]
}

\begin{document}

\maketitle
\thispagestyle{empty}
\pagestyle{empty}

\begin{abstract}
Persistent semantic occupancy mapping is essential for embodied scene understanding. However, perspective-based systems provide limited spatial coverage, while existing panoramic methods primarily predict local volumes from single observations. We introduce PanOVOcc, a training-free framework for persistent open-vocabulary semantic occupancy mapping from panoramic sequences. PanOVOcc unifies panoramic SLAM, open-vocabulary perception, and long-term spatial voxel memory within an online architecture, continuously integrating geometric and semantic evidence into a global, language-queryable map. To facilitate systematic evaluation of this setting, we establish Pan-Replica and Pan-Holo360D, two benchmarks pairing continuous panoramic RGB-D sequences with scene-level semantic occupancy ground truth across synthetic and real-world scenes. Compared with the strongest evaluated baseline for each metric, PanOVOcc improves occupancy IoU and semantic mIoU by absolute $+20.03$ and $+7.06$ on Pan-Replica, and by $+43.26$ and $+20.16$ on Pan-Holo360D, respectively. The source code and the established benchmarks will be available at \url{https://github.com/bakereet/PanOVOcc}.
\end{abstract}

\section{Introduction}
Embodied agents operating in unfamiliar indoor environments require a persistent three-dimensional memory to support navigation, exploration, and interaction. 
Unlike surface reconstruction, semantic occupancy mapping jointly represents occupied structures, traversable free space, unknown regions, and object semantics, providing a direct interface between visual perception and embodied decision-making~\cite{wu2025embodiedocc,jiang2026freeocc,zhu2026gemocc}. 
Recent methods have consequently moved beyond isolated local prediction toward progressively updating a global spatial memory from sequential observations~\cite{wang2025embodiedocc++,wang2026veocc}. 
Nevertheless, most existing online systems are developed primarily for pinhole-camera streams. Due to their limited field of view, individual observations expose only a narrow sector of the environment, resulting in directional blind spots and sparse spatial evidence during exploration.

Panoramic imaging alleviates this limitation by capturing nearly the entire surroundings in a single shot, offering denser spatial evidence around corners and through narrow passages. Recent panorama-based occupancy methods have demonstrated the value of such observations for closed-set semantic occupancy prediction~\cite{shi2026oneocc,zhao2026panommocc} and open-vocabulary occupancy inference~\cite{duan2026o3n}. Yet these methods primarily predict a bounded local occupancy volume from an individual panorama, and do not maintain a persistent spatial memory as the agent moves through the environment. 

A practical embodied map should further support task-dependent semantic queries beyond fixed categories to accommodate long-tailed indoor objects. Open-vocabulary mapping enables this by transferring language-aligned knowledge from pretrained vision-language models into three-dimensional representations~\cite{peng2023openscene,huang2023vlmaps,jatavallabhula2023conceptfusion,yamazaki2024openfusion}.
As illustrated in Fig.~\ref{fig:teaser}, our work enables an embodied agent to continuously accumulate panoramic observations into a global map that can be queried according to downstream task requirements. However, ensuring global consistency remains challenging. 
Sequential data contain both cross-view inconsistencies and substantial observational redundancy, while geometric or semantic errors in individual observations can accumulate over time, degrading the accuracy of the resulting global map~\cite{ma2017multi,marques2024overconfidence,dai2017bundlefusion,morilla2023robust}.
To address these challenges, we propose \textbf{PanOVOcc}, a training-free framework for persistent open-vocabulary semantic occupancy mapping from sequential panoramic observations. 
PanOVOcc integrates panoramic SLAM, open-vocabulary semantic segmentation, and voxel-based mapping into a unified online pipeline. 
As the agent explores, panoramic observations are continuously fused into a long-term spatial voxel memory that maintains a global scene representation. Geometry and language-aligned semantics are incrementally accumulated in this memory, enabling open-vocabulary queries over the evolving 3D environment.
For systematic evaluation, we further introduce \textbf{Pan-Replica} and \textbf{Pan-Holo360D}, two indoor semantic occupancy benchmarks built upon Replica~\cite{straub2019replica} and Holo360D~\cite{ou2026holo360d}, combining continuous panoramic RGB-D sequences with scene-level semantic occupancy annotations across synthetic and real-world environments.
Pan-Replica provides a controlled synthetic benchmark with high-fidelity geometry and dense semantic annotations, enabling systematic evaluation under panoramic observations.
Compared to synthetic environments, Pan-Holo360D offers larger spatial extents, multiple interconnected rooms, and more realistic sensing variations, posing greater challenges for long-term occupancy mapping.

Comprehensive experiments on the Pan-Replica and Pan-Holo360D benchmarks demonstrate the effectiveness of the proposed method across synthetic and real-world settings.
Notably, PanOVOcc achieves substantial gains under the shared eight-category evaluation, with particularly pronounced improvements on large structural categories.
Specifically, the IoUs of ceiling and floor on Pan-Holo360D improve by $37.97$ and $36.23$, respectively.
Beyond the shared taxonomy, it also outperforms FreeOcc~\cite{jiang2026freeocc} on several additional open-vocabulary categories.
Qualitative comparisons further show that our method reconstructs more complete and coherent geometry in large multi-room scenes.

Our main contributions are summarized as follows:

\begin{itemize}
\item We introduce \textbf{PanOVOcc}, a training-free framework for persistent global open-vocabulary semantic occupancy mapping from panoramic sequences, combining omnidirectional sensing with long-term spatial memory for embodied exploration.
\item We develop a geometry-aware persistent fusion scheme that integrates panoramic surface and free-space evidence with reliability-weighted multi-view semantics, while supporting reversible observation-level updates under pose refinement.
\item We introduce \textbf{Pan-Replica} and \textbf{Pan-Holo360D}, two continuous panoramic indoor semantic occupancy benchmarks spanning synthetic and real-world domains.
\end{itemize}

\section{Related Work}
\subsection{Panoramic Semantic Scene Understanding}
Panoramic sensing~\cite{gao2022review} provides complete 360{\textdegree} observations and broad spatial context, making it particularly suitable for embodied scene understanding. 
Existing studies have first established strong image-space semantics by addressing equirectangular projection distortion~\cite{guerrero2020object_recognition,cao2024geometric,zheng2023complementary} and extending panoramic segmentation toward foundation-model-based and open-vocabulary recognition~\cite{zheng2024open,zhong2025omnisam,jiang2026augmenting}.
Beyond image-space perception, increasing attention has been devoted to recovering 3D spatial information from panoramic observations, including depth and structural reconstruction~\cite{wang2020bifuse,wang2022bifuse++,sun2021hohonet}, 3D semantic understanding~\cite{wu2023_3d_segmenter}, and scene reconstruction~\cite{guo2026panovggt,lee2024odgs,spiss2026odgsslam,qiu2026pano3dcomposer}.
More recently, panoramic perception has been extended to voxel-level scene representations for semantic occupancy prediction~\cite{shi2026oneocc,zhao2026panommocc} and open-vocabulary occupancy reasoning~\cite{duan2026o3n}. 
Despite these advances, existing methods mainly focus on reconstructing or predicting local 3D scene content, while persistent semantic mapping from long-horizon panoramic sequences remains underexplored.

\subsection{Embodied Open-Vocabulary Occupancy Mapping}
Occupancy maps explicitly represent occupied, free, and unknown space, providing a natural spatial representation for embodied navigation and exploration~\cite{hornung2013octomap}. 
Early incremental systems explored the fusion of sequential observations into global geometric and semantic representations~\cite{wu2020scfusion}, while recent embodied occupancy methods extend this setting to progressive exploration, continuously updating scene occupancy as an agent observes an initially unknown environment~\cite{wu2025embodiedocc,zhu2026gemocc,wang2025embodiedocc++,wang2026veocc,wang2024embodiedscan}.
However, these methods largely rely on predefined semantic categories. In parallel, open-vocabulary 3D methods inject language-aligned features into scene representations for flexible querying~\cite{peng2023openscene,huang2023vlmaps,jatavallabhula2023conceptfusion,yamazaki2024openfusion,gu2024conceptgraphs}.
Recent work extends this capability to occupancy prediction and mapping~\cite{jiang2026freeocc,zhou2026monocular}, with training-free formulations avoiding task-specific 3D supervision and fixed taxonomies.
Despite these advances, existing methods largely rely on perspective cameras with limited fields of view, requiring more viewpoints and fragmented-view integration to recover complete scene context.
PanOVOcc instead explores panoramic sequences, incrementally integrating observations into a persistent, language-queryable voxel map.

\section{Method}

\subsection{Problem Formulation}

Given a sequential stream of panoramic RGB-D observations
$\mathcal{O}_{1:T}={(\mathbf{I}_t,\mathbf{D}_t)}_{t=1}^{T},$
where $\mathbf{I}_t\in\mathbb{R}^{H\times W\times 3}$ and $\mathbf{D}_t\in\mathbb{R}^{H\times W}$ denote the RGB image and its depth at time $t$, together with an open-vocabulary category set $\mathcal{C}=\{c_1,c_2,\ldots,c_C\}$, our goal is to incrementally construct a persistent 3D semantic occupancy map as an embodied agent explores an initially unknown environment.
At each time step, the map is updated as
\begin{equation}
\mathcal{M}_t=\mathcal{F}\left(\mathcal{M}_{t-1},\mathbf{I}_t,\mathbf{D}_t,\mathcal{C}\right),
\end{equation}
where $\mathcal{M}_t$ is defined in a global coordinate frame and explicitly models the occupancy state of each voxel as occupied, free, or unobserved, together with its accumulated semantic evidence with respect to $\mathcal{C}$.

\begin{figure*}[t]
    \centering

    \includegraphics[width=0.95\textwidth]{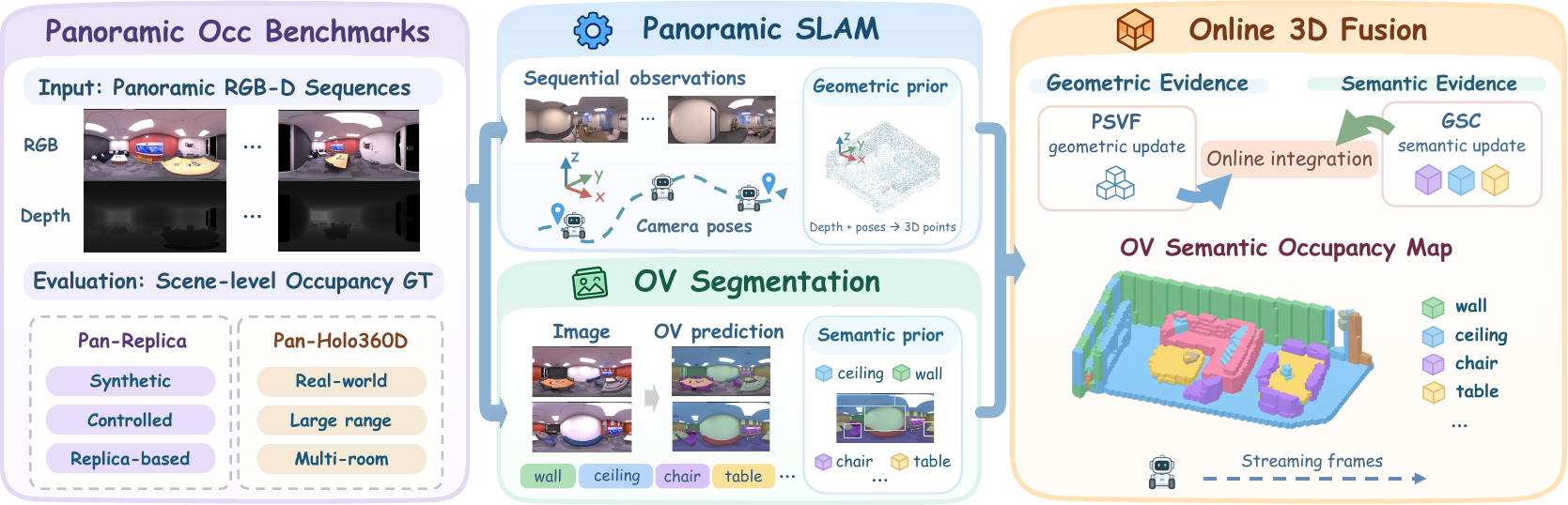}
    \vskip-1ex
    \caption{\textbf{Overview of the benchmarks and PanOVOcc framework.}
    Pan-Replica and Pan-Holo360D provide panoramic RGB-D sequences as mapping inputs and scene-level semantic occupancy ground truth for evaluation.
    Given panoramic RGB-D sequences, PanOVOcc obtains geometric priors from SLAM-estimated camera poses, and semantic priors from open-vocabulary segmentation.
    These priors are incrementally integrated through online 3D fusion to construct a persistent open-vocabulary semantic occupancy map.
    \textit{Abbreviations:}
    Occ: Occupancy;
    OV: Open-Vocabulary;
    GT: Ground Truth.
    }
    \label{fig:pipeline}
    \vskip-3ex
\end{figure*}

\subsection{Overview of PanOVOcc}

As illustrated in Fig.~\ref{fig:pipeline}, PanOVOcc is a training-free framework for online open-vocabulary occupancy mapping from native panoramic sequences. Specifically, it consists of three main components. First, a panoramic SLAM backbone estimates camera poses in a global coordinate frame for map updates. 
Second, open-vocabulary segmentation operates in parallel to provide pixel-wise semantic priors for the input vocabulary.
Third, an online fusion module integrates geometric and semantic evidence into a shared voxel memory. 
It accumulates surface and free-space evidence for occupancy estimation and consolidates multi-view semantics according to prediction reliability and geometric compatibility, and the memory is incrementally updated with reversible observation-level corrections. 
Together, these components continuously transform streaming panoramic observations into a globally aligned, language-queryable semantic occupancy map without task-specific 3D training.

\subsection{Panoramic Geometric Mapping}
\label{sec:geometry}

\noindent\textbf{Panoramic Spatial Representation.}
To incrementally fuse sequential panoramic observations into a global map, we employ ODGS-SLAM~\cite{spiss2026odgsslam} as the pose-estimation backbone with native panoramic support.
For estimated camera pose $\mathbf T_t=(\mathbf R_t,\mathbf t_t)$ at observation $t$, a valid pixel $p=(u,v)$ with radial depth $r_{t,p}$ and unit spherical direction $\mathbf d_p$ is lifted as $\mathbf x_{t,p}=\mathbf R_t(r_{t,p}\mathbf d_p)+\mathbf t_t$.
We maintain a sparse voxel lattice with voxel size $h$, where each voxel $i\in\mathbb Z^3$ is centered at $\mathbf x_i$ and its persistent geometric state is represented by 
\begin{equation}
    \mathcal V_i=(A_i,N_i,F_i,O_i),
    \label{eq:memory_state}
\end{equation}
where $A_i$ accumulates surface support, $N_i$ counts the distinct retained observations contributing to that support, $F_i$ accumulates free-space evidence, and $O_i$ records valid depth-comparison evidence.
These quantities are stored in a spatially hashed block structure, and semantic evidence is associated with the same lattice as described in Sec.~\ref{sec:semantic}.

\vspace{3pt}\noindent\textbf{Panoramic Signed Voxel Fusion.}
ERP measurements exhibit nonuniform spatial sampling, with tangential footprints varying by latitude and range. We model each observation using an anisotropic kernel aligned with its spherical viewing geometry.
For latitude $\phi$, the longitudinal, latitudinal, and radial scales are defined as
{\setlength{\abovedisplayskip}{5pt}
 \setlength{\belowdisplayskip}{5pt}
\begin{equation}
    \boldsymbol{\sigma}_{t,p} = \left[ r|\cos\phi|\Delta\theta,\; r\Delta\phi,\; \alpha r+\beta \right].
    \label{eq:psvf_scales}
\end{equation}
}

For sampling stride $s$, the angular intervals are $\Delta\theta=2\pi s/W$ and $\Delta\phi=\pi s/H$. The parameters $\alpha$ and $\beta$ model range-dependent radial uncertainty. Let $\mathbf E_{t,p}$ contain orthonormal longitude, latitude, and ray directions transformed into the mapping frame. The corresponding
covariance is $\boldsymbol\Sigma_{t,p}=\mathbf E_{t,p}\operatorname{diag}(\boldsymbol\sigma^2_{t,p})\mathbf E_{t,p}^{\top}$, and the squared Mahalanobis distance is $d_{t,p}^2(i)=(\mathbf x_i-\mathbf x_{t,p})^{\top}\boldsymbol\Sigma_{t,p}^{-1}(\mathbf x_i-\mathbf x_{t,p})$.

Surface contributions are aggregated within each observation before being
written to the persistent map:
{\setlength{\abovedisplayskip}{5pt}
 \setlength{\belowdisplayskip}{5pt}
\begin{equation}
    \Delta A_i^{(t)}
    =\sum_{p\in\mathcal P_t(i)}
       \omega(\phi,r)
       \exp\!\left(-\tfrac12 d_{t,p}^2(i)\right)
       \mathbb I\!\left[d_{t,p}^2(i)\leq\kappa^2\right].
    \label{eq:memory_surface_delta}
\end{equation}
}
Here, $\mathcal P_t(i)$ contains valid strided measurements whose bounded neighborhoods include voxel $i$, and $\mathbb I[\cdot]$ is the indicator function. The weight $\omega$ models latitude-dependent sampling reliability and range-dependent depth uncertainty, and $\kappa$ sets the Mahalanobis truncation radius. The view-support increment is $\Delta N_i^{(t)}=\mathbb I[\Delta A_i^{(t)}>0]$. Thus, multiple pixels from one panorama may increase surface support but contribute at most one view count.

To incorporate free-space evidence, we project active voxel centers into the incoming panorama. Let $r_{i,t}$ denote the camera-to-voxel range and $\widehat r_{i,t}$ the observed radial depth at the projected pixel; each valid depth comparison contributes $\Delta O_i^{(t)}$, and it additionally contributes $\Delta F_i^{(t)}$ when the local depth neighborhood is stable and $r_{i,t}<\widehat r_{i,t}-\epsilon(\widehat r_{i,t})$, where $\epsilon(r)=\epsilon_{\mathrm{abs}}+\epsilon_{\mathrm{rel}}r$ accounts for absolute and range-dependent depth uncertainty. The accumulated geometric evidence yields the signed score
\begin{equation}
    g_i=\frac{A_i}{\max\{N_i,1\}}
          -\lambda_f\frac{F_i}{\max\{O_i,1\}},
    \label{eq:memory_signed_score}
\end{equation}
where $\lambda_f$ controls the contribution of free-space evidence.
The occupied set $\Omega$ is determined from $g_i$.
Given a strong surface threshold $\tau_s$, a weaker threshold $\tau_w$, and a maximum admissible free-space ratio $\gamma_f$, strong supported surfaces are accepted directly, whereas weaker surface responses are retained only if the free-space ratio does not exceed $\gamma_f$.

\noindent\textbf{Reversible Persistent Update.}
Pose refinement can change the spatial association of previously integrated measurements. RPU retains each observation's geometric contribution as a sparse delta $\Delta\mathcal V_t$, containing the voxel-indexed increments
$(\Delta A_i^{(t)},\Delta N_i^{(t)},\Delta F_i^{(t)},\Delta O_i^{(t)})$.
For the currently retained observation set $\mathcal K$, the geometric memory is represented by
$\mathcal V=\sum_{t\in\mathcal K}\Delta\mathcal V_t$.
When the pose of observation $t$ is revised to $\mathbf T'_t$,
its stored contribution is removed and recomputed under the revised pose:
{\setlength{\abovedisplayskip}{5pt}
 \setlength{\belowdisplayskip}{5pt}
\begin{equation}
    \mathcal V\leftarrow
    \mathcal V-\Delta\mathcal V_t^{\mathrm{old}}
                +\Delta\mathcal V_t^{\mathrm{new}}(\mathbf T'_t).
    \label{eq:memory_reversible}
\end{equation}
}

Storing raw additive statistics permits these operations before nonlinear occupancy decisions while preserving provenance for revising individual observations. The same principle extends to semantic evidence, as detailed below.

\begin{table}[t]
\centering
\caption{
\textbf{Comparison with representative embodied and panoramic occupancy benchmarks.}
\perspicon{} / \panoicon{}: perspective / panoramic observations.
In. / Out.: Indoor / Outdoor.
}
\vskip -1ex
\label{tab:benchmark_comparison}
\scriptsize
\setlength{\tabcolsep}{3.0pt}
\renewcommand{\arraystretch}{1.08}
\begin{tabular}{@{}lcccc@{}}
\toprule
\textbf{Benchmark} &
\textbf{Obs.} &
\textbf{Temporal} &
\textbf{Scope} &
\textbf{Domain}\\
\midrule

EmbodiedOcc-ScanNet~\cite{wu2025embodiedocc}
& \perspicon{}
& Sequence
& Scene
& In.-Real\\

ReplicaOcc~\cite{jiang2026freeocc}
& \perspicon{}
& Sequence
& Scene
& In.-Synth.\\

QuadOcc~\cite{shi2026oneocc}
& \panoicon{}
& Single
& Local
& Out.-Real\\

Human360Occ~\cite{shi2026oneocc}
& \panoicon{}
& Single
& Local
& Out.-Synth.\\

PanoMMOcc~\cite{zhao2026panommocc}
& \panoicon{}
& Single
& Local
& Out.-Real\\

HIOcc~\cite{zhu2026gemocc}
& \perspicon{}
& Both
& Local / Scene
& In.-Real\\

HIOcc~\cite{zhu2026gemocc}
& \panoicon{}
& Station-based
& Scene
& In.-Real\\

\midrule
\textbf{Pan-Replica}
& \panoicon{}
& \textbf{Sequence}
& \textbf{Scene}
& \textbf{In.-Synth.}\\

\textbf{Pan-Holo360D}
& \panoicon{}
& \textbf{Sequence}
& \textbf{Scene}
& \textbf{In.-Real}\\

\bottomrule
\end{tabular}
\vskip -3ex
\end{table}

\newcommand{\datasetimage}[2][1]{%
    \centering
    \includegraphics[
        width=#1\linewidth
    ]{figures/dataset/Indoor_#2.png}%
    \par
}

\subsection{Geometry-Aware Semantic Consolidation}
\label{sec:semantic}

\noindent\textbf{Multi-Observation Semantic Evidence.}
Given a category set $\mathcal C=\{c_1,\ldots,c_C\}$, we employ Open Panoramic Segmentation~\cite{zheng2024open} to obtain semantic observations directly from native ERP images. 
For observation $t$, the segmenter predicts a semantic logit vector $\boldsymbol{\ell}_t(u,v) \in \mathbb R^C$ for each pixel $(u,v)$, which is converted into a categorical distribution $\mathbf p_t(u,v) = \operatorname{softmax} \left( \boldsymbol{\ell}_t(u,v) \right).$

Since panoramic observations exhibit spatially varying reliability, each semantic measurement is assigned a weight
$\eta_{t,p}=c_{t,p}\,w_{\mathrm{erp}}(\phi)\,w_{\mathrm{range}}(r)$,
where $c_{t,p}$ is the maximum predicted class probability, and the remaining factors account for latitude-dependent and range-dependent reliability. 
Each voxel maintains raw semantic evidence
$E_i\in\mathbb R^C$ and total weight $S_i$.
The per-observation semantic contribution is
{\setlength{\abovedisplayskip}{5pt}
 \setlength{\belowdisplayskip}{5pt}
\begin{equation}
    \begin{bmatrix}\Delta E_i^{(t)}\\\Delta S_i^{(t)}\end{bmatrix}
    =\sum_{\substack{p\in\mathcal P_t}}
       \eta_{t,p}
       \begin{bmatrix}\mathbf p_t(p)\\1\end{bmatrix}.
    \label{eq:memory_semantic_delta}
\end{equation}
}

\noindent\textbf{Geometry-Aware Semantic Consolidation.}
Although multi-observation accumulation reduces temporal variability, ambiguous voxels may still remain in regions with inconsistent segmentation. 
We regularize these unaries on a sparse graph $\mathcal G=(\Omega,\mathcal E)$ whose vertices are occupied voxels. Candidate edges connect six-neighbor voxel pairs, restricting graph propagation to the reconstructed occupied support. For adjacent voxels, the affinity is defined as
\begin{equation}
    W_{ij}=\kappa_g(i,j)\,\kappa_n(i,j)\,
           \kappa_t(i,j)\,\kappa_c(i,j),
    \label{eq:memory_gsc_affinity}
\end{equation}
where the four terms encode geometric and appearance cues. Specifically, $\kappa_g$ measures the consistency of the signed PSVF scores using an exponential penalty on $|g_i-g_j|$; $\kappa_n$ favors aligned surface normals; $\kappa_t$ suppresses connections along the local surface-normal direction and therefore encourages propagation within the tangent plane; and $\kappa_c$ measures similarity between the accumulated mean RGB colors.
The weights are row-normalized as $\bar W_{ij}=W_{ij}/\sum_{l\in\mathcal N(i)}W_{il}$, where $\mathcal N(i)$ is the candidate neighborhood.
Starting from $\mathbf q_i^{(0)}=\mathbf p_i$, GSC performs anchored diffusion:
\begin{equation}
    \mathbf q_i^{(k+1)}
    =\frac{\rho_i\mathbf p_i+
       \lambda_s\displaystyle\sum_{j\in\mathcal N(i)}
       \bar W_{ij}\mathbf q_j^{(k)}}
       {\rho_i+\lambda_s\displaystyle\sum_{j\in\mathcal N(i)}\bar W_{ij}},
    \label{eq:memory_gsc_update}
\end{equation}
where $\lambda_s$ controls spatial regularization and $\rho_i$ is an anchor strength derived from accumulated semantic weight and normalized entropy. Reliable observations therefore provide stronger anchors, whereas uncertain voxels receive greater relative influence from geometrically compatible neighbors.

\vspace{3pt}\noindent\textbf{Persistent Semantic Update.}
The semantic memory follows the same observation-level decomposition as the geometric memory. Each frame retains a sparse semantic delta $\Delta\mathcal S_t=(\Delta E_t,\Delta S_t)$ together with the RGB, normal, and observation-count statistics used by GSC. 
When a pose revision changes voxel associations, the previous semantic contribution is subtracted, and a new contribution is integrated using the revised pose, following
Eq.~\eqref{eq:memory_reversible}.

The output assigns the maximum-probability semantic category only to occupied voxels. A bounded nearest-evidence transfer supplies labels where direct semantic support is absent; this output operation is shared by the raw and GSC variants. Occupied voxels without eligible semantic evidence remain semantically unknown.

\section{Experiments}

\begin{figure}[t]
    \centering

    \begin{minipage}[c]{0.26\linewidth}
        \datasetimage[1.15]{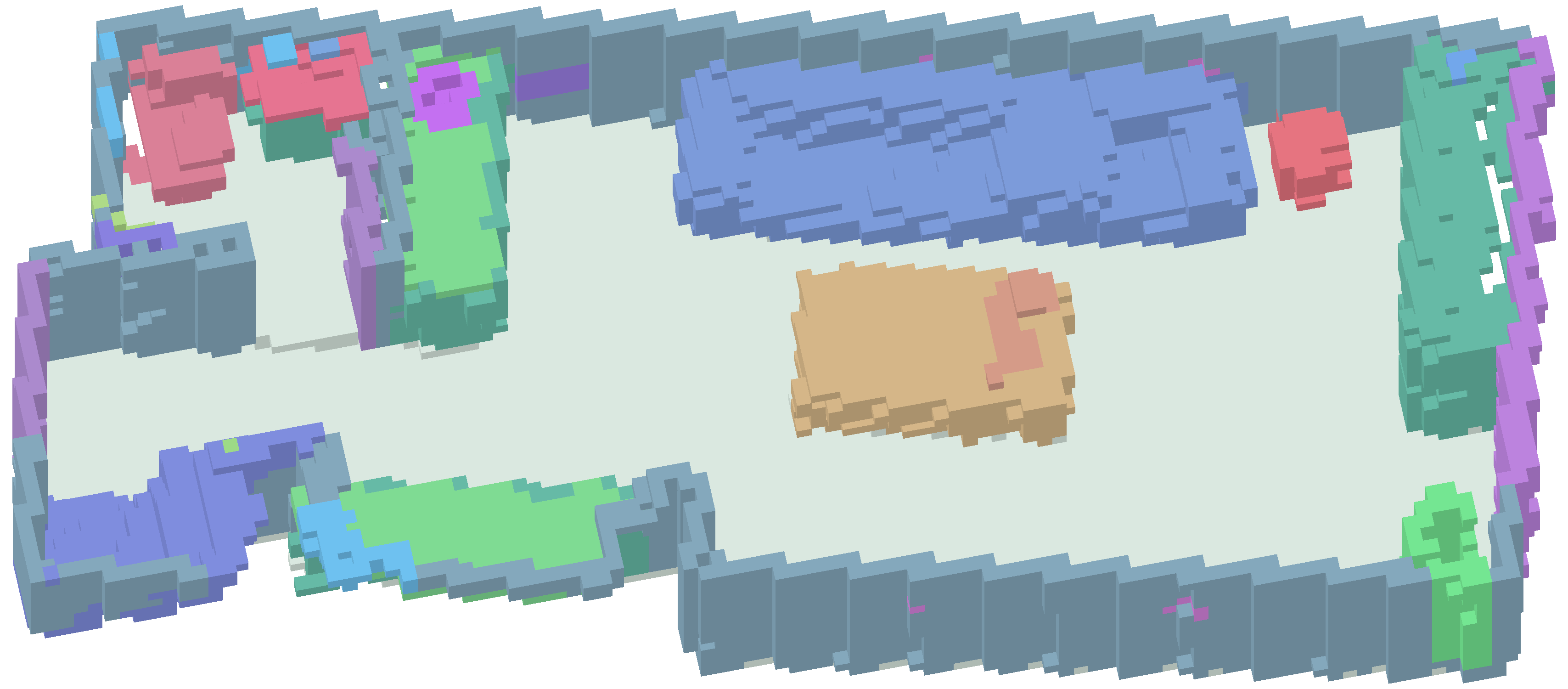}

        \vspace{6pt}  

        \datasetimage[1.15]{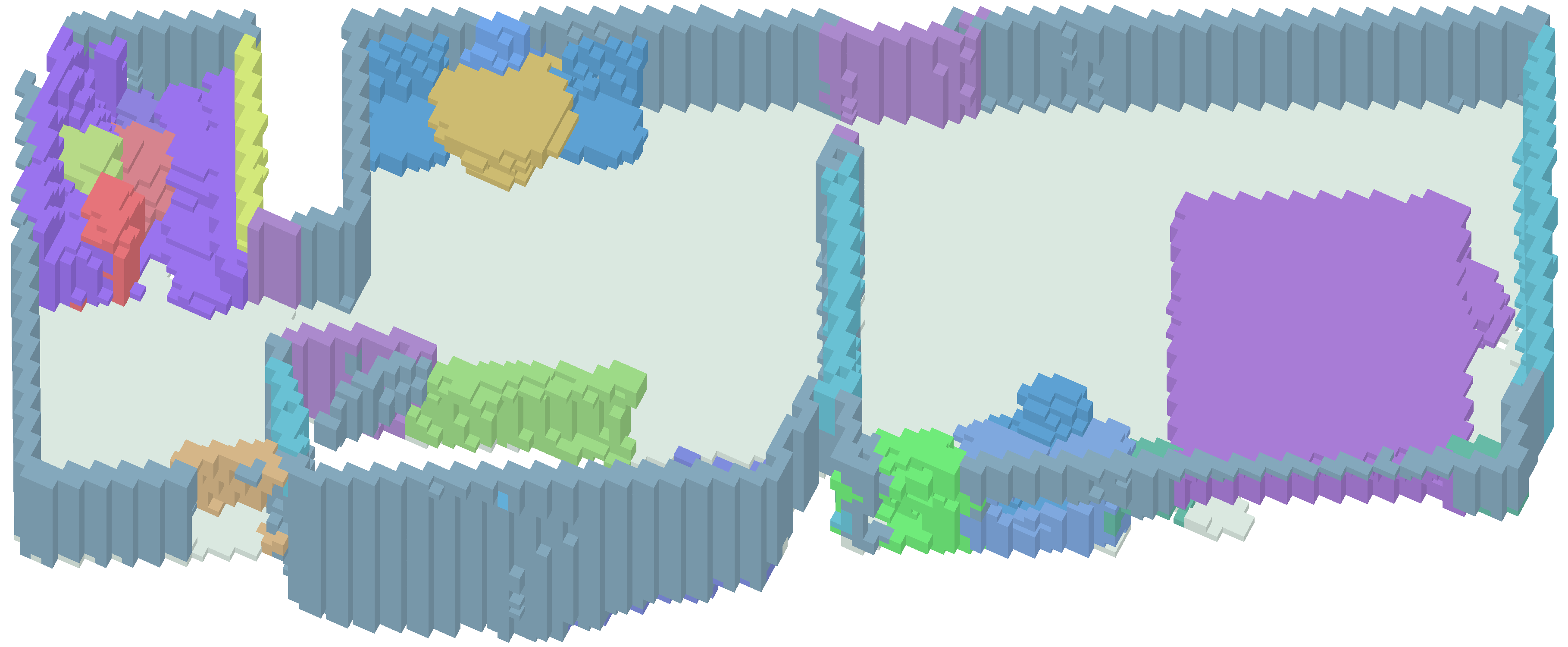}
    \end{minipage}
    \hspace{6pt}%
    \begin{minipage}[c]{0.35\linewidth}
        \datasetimage[0.7]{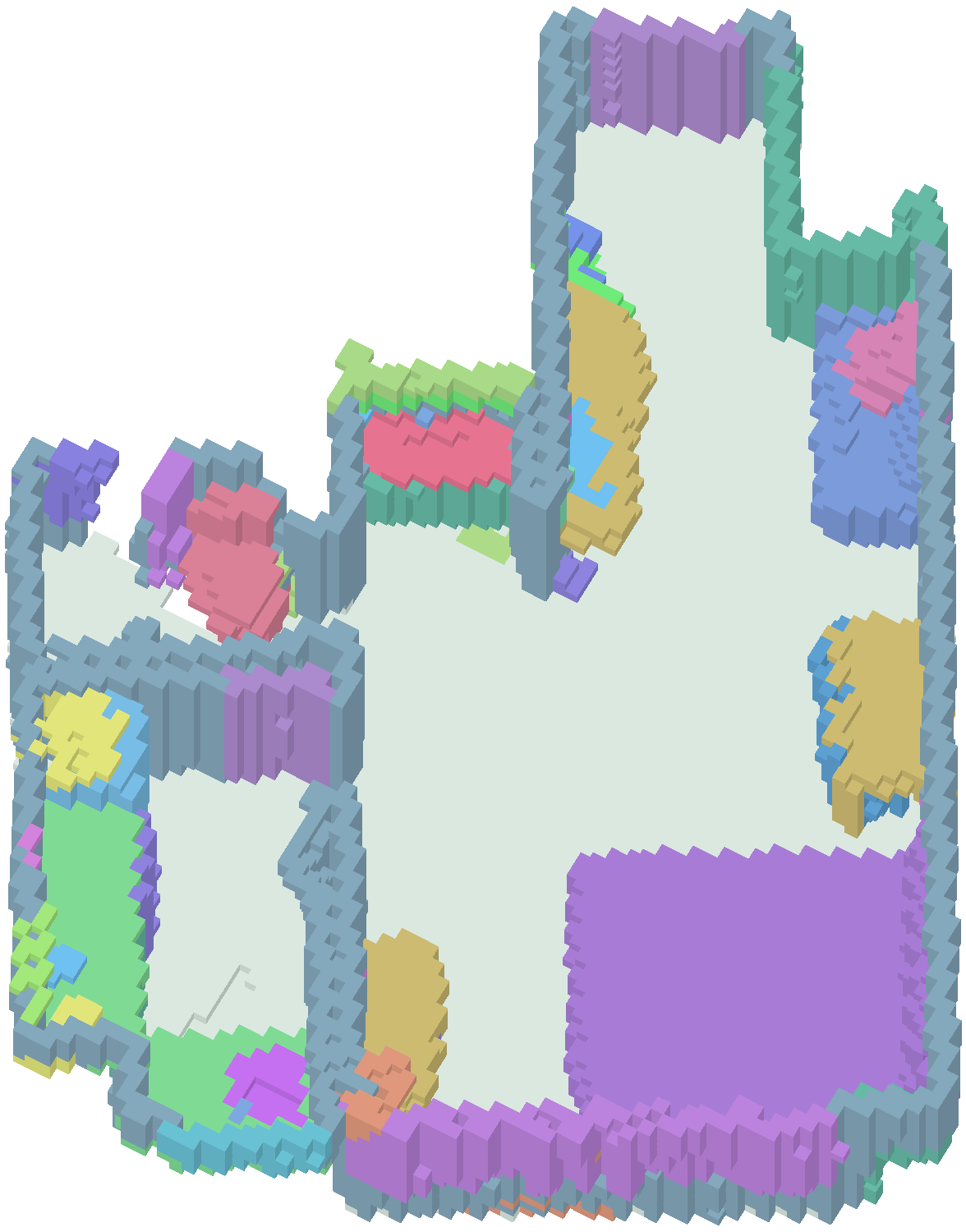}
    \end{minipage}\hfill%
    \begin{minipage}[c]{0.35\linewidth}
        \datasetimage[0.75]{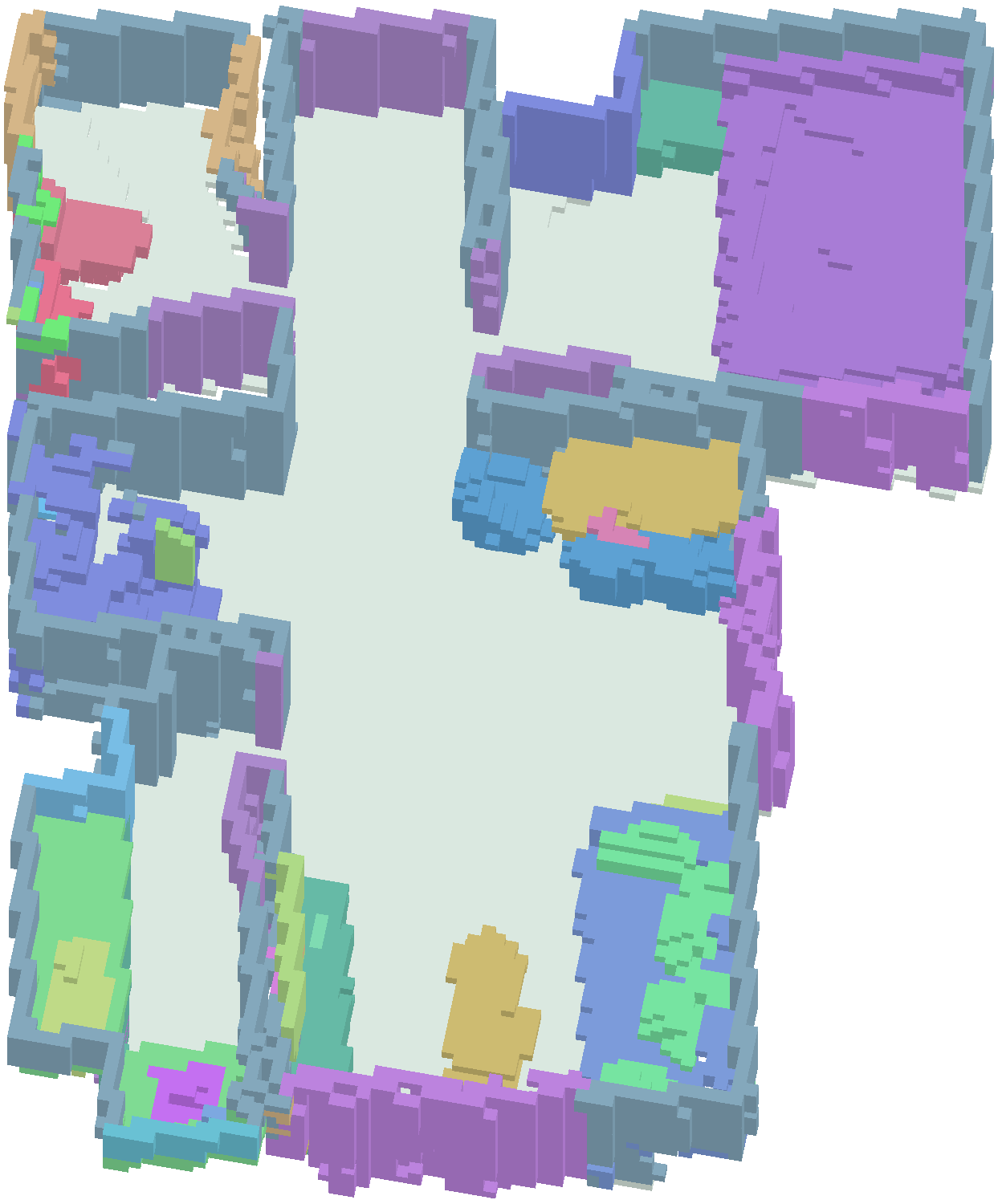}
    \end{minipage}

    \par\vspace{4pt} 

    \begin{minipage}[c]{0.32\linewidth}
        \datasetimage[0.95]{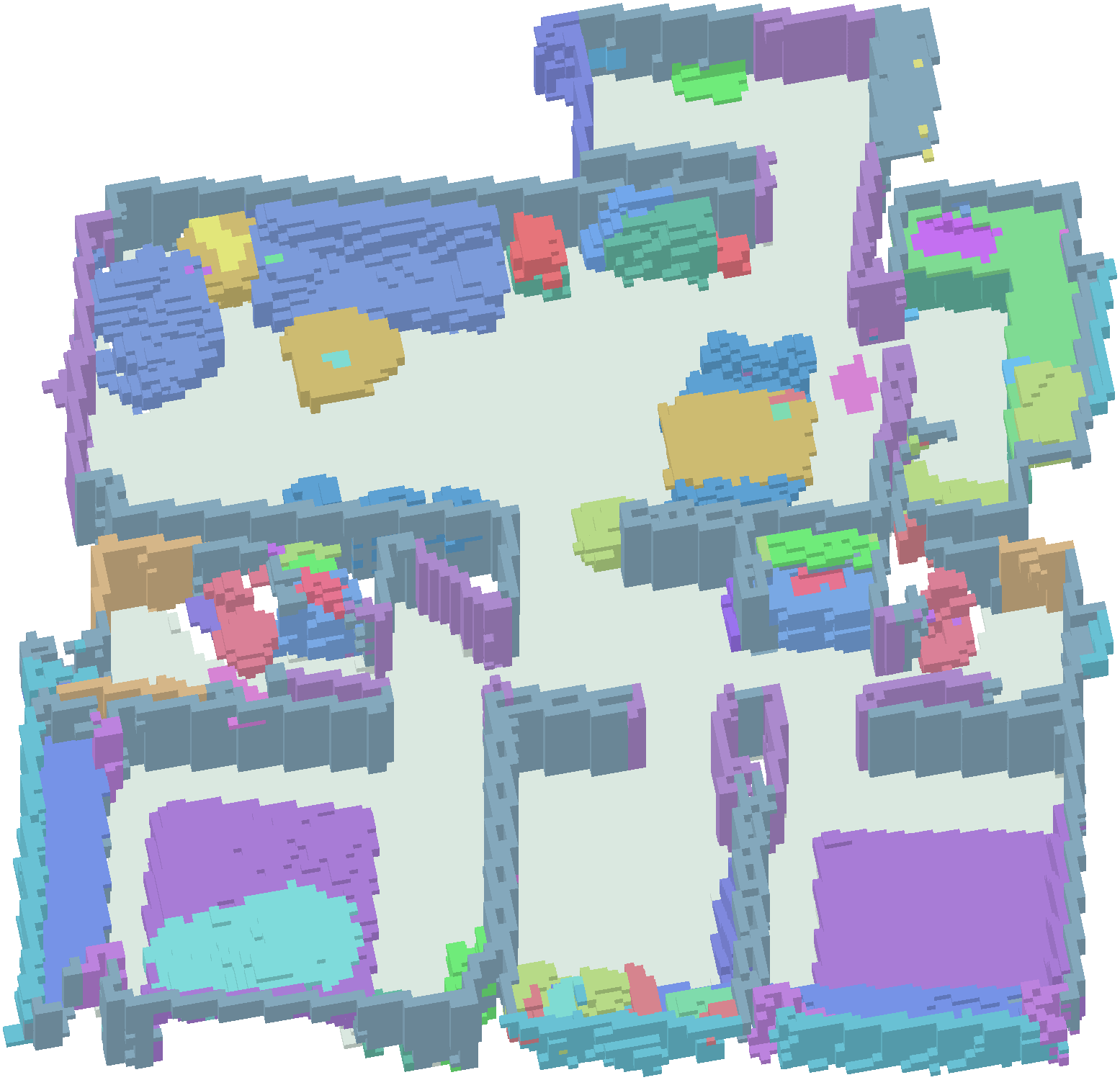}
    \end{minipage}\hfill%
    \begin{minipage}[c]{0.32\linewidth}
        \datasetimage[0.85]{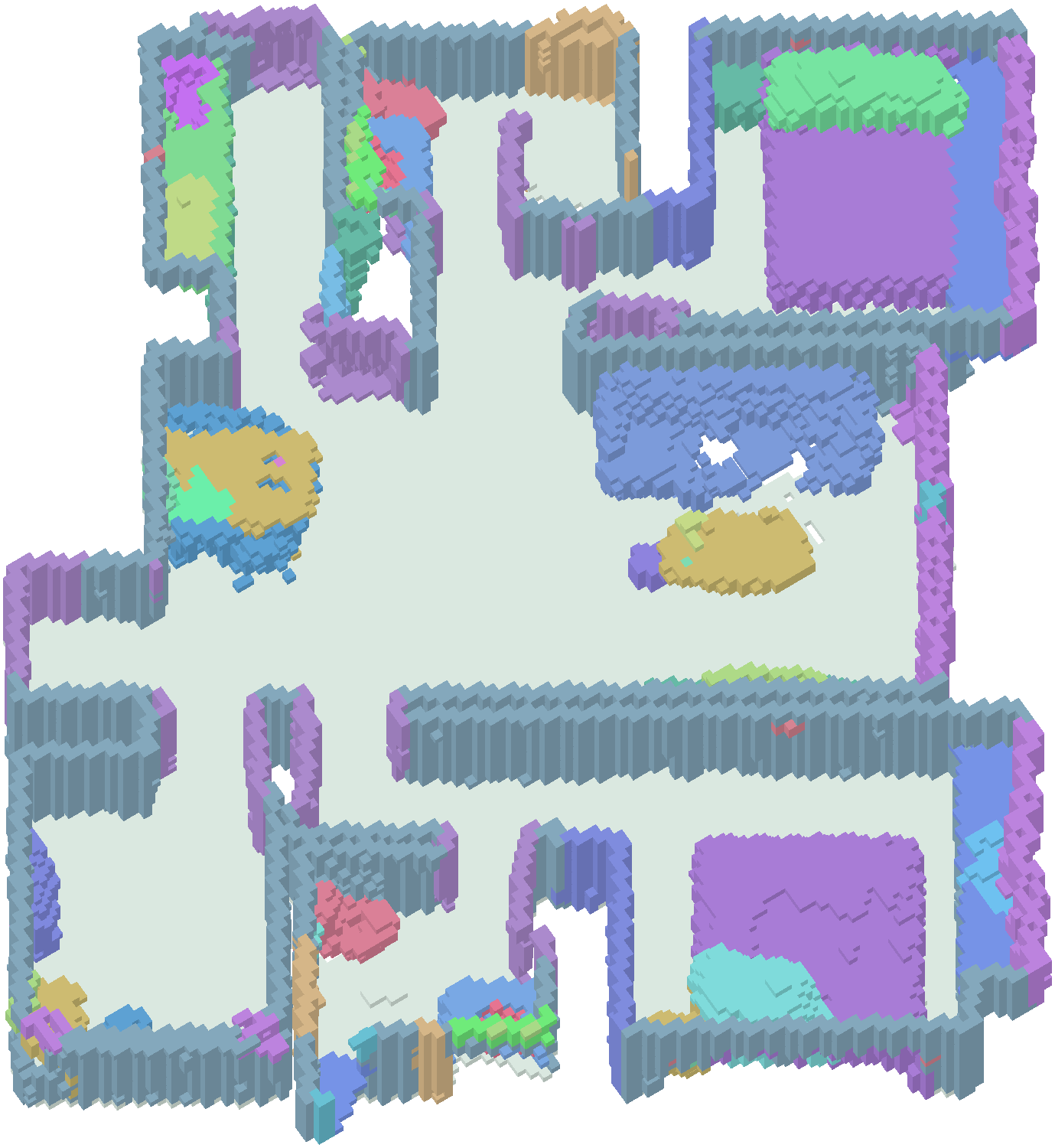}
    \end{minipage}\hfill%
    \begin{minipage}[c]{0.32\linewidth}
        \datasetimage[0.85]{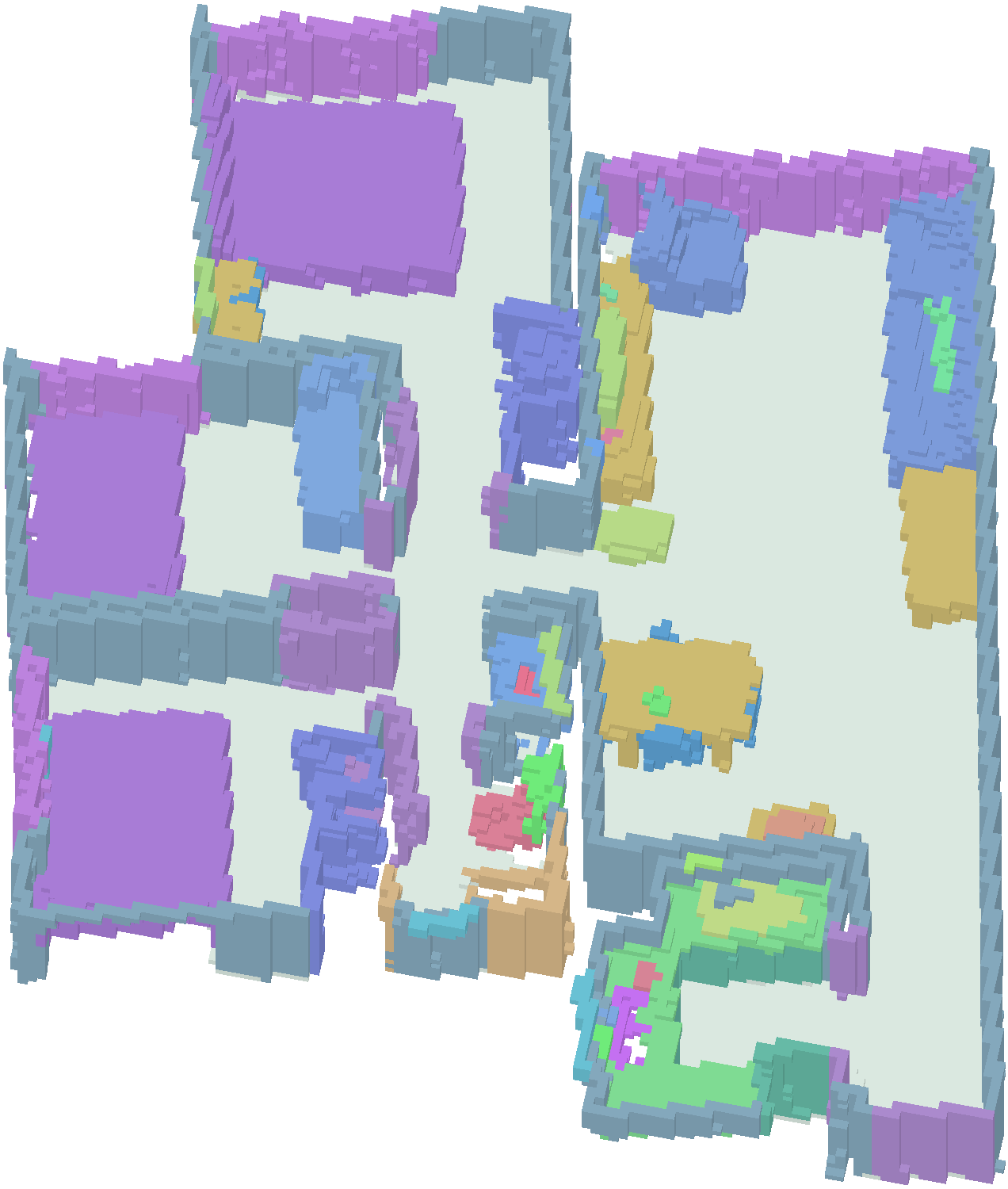}
    \end{minipage}

    \caption{\textbf{Overview of Pan-Holo360D.}
    Scene-level semantic occupancy GT for seven real-world indoor scenes selected from Holo360D~\cite{ou2026holo360d}.}
    \label{fig:dataset_overview}
    \vskip -3ex
\end{figure}

\begin{table*}[!t]
    \centering
    \caption{
        \textbf{Quantitative comparison on Pan-Replica.}
        Perspective methods receive twelve directional views generated from
        each panoramic observation, providing approximately $360^\circ$ horizontal coverage.
    }
    \label{tab:panreplica_main}

    \vskip-1ex

    \begin{tabular}{
        l
        @{\hspace{7pt}}
        ccc
        @{\hspace{9pt}}
        cc
        @{\hspace{9pt}}
        cccccccc
    }
        \toprule

        \textbf{Method}
        &
        \rotatebox{90}{\textbf{Obs.}}
        &
        \rotatebox{90}{\textbf{Anno.}}
        &
        \rotatebox{90}{\textbf{Pose}}
        &
        \textbf{IoU}
        &
        \textbf{mIoU}
        &
        \rotatebox{90}{\textcolor{ceilingcolor}{$\blacksquare$} ceiling}
        &
        \rotatebox{90}{\textcolor{floorcolor}{$\blacksquare$} floor}
        &
        \rotatebox{90}{\textcolor{wallcolor}{$\blacksquare$} wall}
        &
        \rotatebox{90}{\textcolor{windowcolor}{$\blacksquare$} wind.}
        &
        \rotatebox{90}{\textcolor{chaircolor}{$\blacksquare$} chair}
        &
        \rotatebox{90}{\textcolor{bedcolor}{$\blacksquare$} bed}
        &
        \rotatebox{90}{\textcolor{sofacolor}{$\blacksquare$} sofa}
        &
        \rotatebox{90}{\textcolor{tablecolor}{$\blacksquare$} table}
        \\

        \midrule

        \multicolumn{14}{l}{
            \textit{\color{gray!70!black}Supervised learning}
        }
        \\[-1pt]

        EmbodiedOcc~\cite{wu2025embodiedocc}
        &
        \perspicon{}
        &
        \cmark
        &
        \cmark
        &
        18.02 & 5.58
        &
        3.19
        &
        6.34
        &
        10.05
        &
        3.44
        &
        4.26
        &
        7.78
        &
        6.81
        &
        2.76
        \\

        EmbodiedOcc++~\cite{wang2025embodiedocc++}
        &
        \perspicon{}
        &
        \cmark
        &
        \cmark
        &
        19.41 & 5.96
        &
        3.04
        &
        7.89
        &
        10.61
        &
        4.26
        &
        4.31
        &
        6.34
        &
        7.06
        &
        4.18
        \\

        VEOcc~\cite{wang2026veocc}
        &
        \perspicon{}
        &
        \cmark
        &
        \cmark
        &
        20.83 & 17.02
        &
        19.08
        &
        12.42
        &
        13.83
        &
        7.96
        &
        \textbf{21.26}
        &
        8.96
        &
        34.29
        &
        \underline{18.41}
        \\

        GPOcc~\cite{zhou2026gpocc}
        &
        \perspicon{}
        &
        \cmark
        &
        \cmark
        &
        19.64 & 15.72
        &
        21.14
        &
        19.98
        &
        9.75
        &
        7.22
        &
        14.53
        &
        10.96
        &
        29.87
        &
        12.30
        \\

        \midrule

        \multicolumn{14}{l}{
            \textit{\color{gray!70!black}Training-free}
        }
        \\[-1pt]

        FreeOcc (RGBD)~\cite{jiang2026freeocc}
        &
        \perspicon{}
        &
        \xmark
        &
        \xmark
        &
        \underline{55.60} & \underline{29.93}
        &
        \underline{39.57}
        &
        \underline{41.83}
        &
        \underline{39.77}
        &
        \underline{20.55}
        &
        \underline{17.43}
        &
        \textbf{29.81}
        &
        \underline{38.05}
        &
        12.45
        \\

        \addlinespace[1pt]

        \textbf{PanOVOcc (Ours)}
        &
        \panoicon{}
        &
        \xmark
        &
        \xmark
        &
        \textbf{75.63}
        &
        \textbf{36.99}
        &
        \textbf{54.79}
        &
        \textbf{57.26}
        &
        \textbf{58.47}
        &
        \textbf{23.93}
        &
        16.04
        &
        \underline{17.30}
        &
        \textbf{45.25}
        &
        \textbf{22.87}
        \\

        \bottomrule
    \end{tabular}
\vskip -3ex
\end{table*}

In this section, we evaluate PanOVOcc with the aim of answering three questions:

\begin{itemize}
    \item \textbf{Q1:} \textit{How effectively do panoramic sequential observations support scene-level occupancy mapping compared with existing methods?} (Sec.~\ref{sec:quantitative})
    \item \textbf{Q2:} \textit{How well does PanOVOcc generalize from controlled synthetic to real-world panoramic sequences without retraining?} (Sec.~\ref{sec:quantitative})
    \item \textbf{Q3:} \textit{What factors govern the performance of PanOVOcc, particularly with respect to its geometric and semantic components, and the quality of upstream predictions?} (Sec.~\ref{sec:ablation})
\end{itemize}

\subsection{Benchmark}

We evaluate PanOVOcc on two panoramic evaluation sets.
\textbf{Pan-Replica} supports controlled evaluation of panoramic observations in synthetic environments, whereas \textbf{Pan-Holo360D} assesses real-world generalization on native continuous panoramic sequences with newly constructed scene-level semantic occupancy annotations.
Table~\ref{tab:benchmark_comparison} compares our evaluation sets with representative embodied and panoramic occupancy benchmarks.

\vspace{3pt}\noindent\textbf{Pan-Replica.}
We construct Pan-Replica as a panoramic evaluation variant of ReplicaOcc~\cite{jiang2026freeocc}, retaining its camera trajectories~\cite{zhu2022niceslam} and scene-level occupancy ground truth.
Using a rendering pipeline adapted from MatryODShka~\cite{attal2020matryodshka}, we render panoramic RGB-D observations from the Replica assets.
The resulting sequences retain the camera centers and temporal sampling of their perspective counterparts while providing omnidirectional observations at each viewpoint.

\vspace{3pt}\noindent\textbf{Pan-Holo360D.}
We further construct Pan-Holo360D using seven indoor scenes from Holo360D~\cite{ou2026holo360d}, which provide long, continuous panoramic RGB-D sequences with aligned camera poses, exhibiting substantially more realistic sensing noise and appearance variation than synthetic scenes.
Following the scene-level occupancy construction protocol of ReplicaOcc~\cite{jiang2026freeocc}, we aggregate depth observations from each complete sequence in the global coordinate system to construct a voxelized scene representation.
Since Holo360D lacks semantic annotations suitable for occupancy evaluation, we manually annotate a sparse set of representative panoramic frames. We then lift the labels to the voxel grid using depth and camera poses, consolidate multi-view observations, and inspect and correct the 3D semantic volume.
As shown in Fig.~\ref{fig:dataset_overview}, the selected sequences span house-scale environments with multiple interconnected rooms.

\subsection{Experimental Setup}
\noindent\textbf{Setup and Metrics.} 
Experiments are conducted on a single NVIDIA RTX 4090 GPU with a voxel resolution of $0.08m$ for occupancy mapping. 
We evaluate on the Pan-Replica and Pan-Holo360D benchmarks using occupancy IoU and semantic mIoU. For compatibility with supervised baselines trained under a fixed semantic taxonomy, the main comparison follows the FreeOcc protocol~\cite{jiang2026freeocc} and evaluates semantic occupancy over the eight shared categories: \emph{ceiling}, \emph{floor}, \emph{wall}, \emph{window}, \emph{chair}, \emph{bed}, \emph{sofa}, and \emph{table}.
For open-vocabulary visualization and supplementary evaluation, the original fine-grained semantic annotations are retained.

\vspace{3pt}\noindent\textbf{Perspective Baseline Adaptation.}
Existing indoor occupancy methods are predominantly designed for perspective image sequences and cannot directly process equirectangular panoramas.
To provide these methods with angular coverage comparable to the panoramic input, each panoramic frame is projected into twelve perspective views, comprising four azimuthal directions at each of the upper, middle, and lower elevation levels. Views with the same orientation form temporally continuous sequences that are processed independently. Predictions are expressed in the dataset’s global coordinate frame using the supplied camera poses; for FreeOcc, each independently estimated SLAM trajectory is aligned to the corresponding reference trajectory before transforming its Gaussian map. For voxel-based outputs, overlapping occupied predictions are combined by union, and semantic conflicts are resolved by class voting for dense outputs and maximum class confidence for sparse outputs.

\definecolor{curtaincolor}{RGB}{188, 189, 34}
\definecolor{doorcolor}{RGB}{140, 86, 75}
\definecolor{tvcolor}{RGB}{31, 119, 180}
\definecolor{lightingcolor}{RGB}{255, 187, 120}

\begin{table*}[!t]
    \centering
    \caption{
        \textbf{Quantitative comparison on Pan-Holo360D.}
        We report semantic occupancy performance on eight categories shared with Pan-Replica and four additional open-vocabulary categories.
    }
    \label{tab:panholo360d_main}

    \vskip-1ex

    \begin{tabular}{
        l
        @{\hspace{7pt}}
        c
        @{\hspace{9pt}}
        cc
        @{\hspace{9pt}}
        cccccccc
        @{\hspace{9pt}}
        cccc
    }
        \toprule

        &
        &
        &
        &
        \multicolumn{8}{c}{\textbf{Overlap-8 Vocabulary}}
        &
        \multicolumn{4}{c}{\textbf{Open-Vocabulary}}
        \\[0pt]

        \cmidrule(lr){5-12}
        \cmidrule(lr){13-16}

        \textbf{Method}
        &
        \rotatebox{90}{\textbf{Obs.}}
        &
        \rotatebox{90}{\textbf{IoU}}
        &
        \rotatebox{90}{\textbf{mIoU}(8)}
        &
        \rotatebox{90}{\textcolor{ceilingcolor}{$\blacksquare$} ceiling}
        &
        \rotatebox{90}{\textcolor{floorcolor}{$\blacksquare$} floor}
        &
        \rotatebox{90}{\textcolor{wallcolor}{$\blacksquare$} wall}
        &
        \rotatebox{90}{\textcolor{windowcolor}{$\blacksquare$} wind.}
        &
        \rotatebox{90}{\textcolor{chaircolor}{$\blacksquare$} chair}
        &
        \rotatebox{90}{\textcolor{bedcolor}{$\blacksquare$} bed}
        &
        \rotatebox{90}{\textcolor{sofacolor}{$\blacksquare$} sofa}
        &
        \rotatebox{90}{\textcolor{tablecolor}{$\blacksquare$} table}
        &
        \rotatebox{90}{\textcolor{curtaincolor}{$\blacksquare$} curtain}
        &
        \rotatebox{90}{\textcolor{doorcolor}{$\blacksquare$} door}
        &
        \rotatebox{90}{\textcolor{tvcolor}{$\blacksquare$} tv}
        &
        \rotatebox{90}{\textcolor{lightingcolor}{$\blacksquare$} lighti.}
        \\

        \midrule

        \multicolumn{16}{l}{
            \textit{\color{gray!70!black}Supervised learning}
        }
        \\[-1pt]

        EmbodiedOcc~\cite{wu2025embodiedocc}
        &
        \perspicon{}
        &
        18.21 & 5.16
        & 0.15
        & 13.35
        & 9.65
        & 4.10
        & 3.08
        & 4.30
        & 3.12
        & 3.49
        &
        \xmarkblack
        & \xmarkblack
        & \xmarkblack
        & \xmarkblack
        \\

        EmbodiedOcc++~\cite{wang2025embodiedocc++}
        &
        \perspicon{}
        &
        18.02 & 4.99
        & 0.16
        & 13.21
        & 10.15
        & 3.07
        & 2.81
        & 3.81
        & 2.95
        & 3.74

        & \xmarkblack
        & \xmarkblack
        & \xmarkblack
        & \xmarkblack
        \\

        VEOcc~\cite{wang2026veocc}
        &
        \perspicon{}
        &
        \underline{23.01} & 12.49
        & 15.64
        & 7.81
        & \underline{18.72}
        & \underline{13.41}
        & \underline{11.47}
        & 4.10
        & 15.85
        & \underline{12.89}
        
        & \xmarkblack
        & \xmarkblack
        & \xmarkblack
        & \xmarkblack
        \\

        GPOcc~\cite{zhou2026gpocc}
        &
        \perspicon{}
        &
        22.03 & \underline{14.56}
        & \underline{29.20}
        & 11.15
        & 17.24
        & 12.88
        & 8.54
        & 8.20
        & \underline{21.51}
        & \underline{7.76}
        
        & \xmarkblack
        & \xmarkblack
        & \xmarkblack
        & \xmarkblack
        \\

        \midrule

        \multicolumn{16}{l}{
            \textit{\color{gray!70!black}Training-free}
        }
        \\[-1pt]

        FreeOcc (RGBD)~\cite{jiang2026freeocc}
        &
        \perspicon{}
        &
        20.66 & 10.01
        & 13.13
        & \underline{13.38}
        & 12.17
        & 7.76
        & 4.89
        & \underline{11.37}
        & 11.74
        & 5.62
        
        & 13.47
        & 3.11
        & 6.40
        & 8.21
        \\

        \addlinespace[1pt]

        \textbf{PanOVOcc (Ours)}
        &
        \panoicon{}
        &
        \textbf{66.27} & \textbf{34.72}
        & \textbf{67.17}
        & \textbf{49.61}
        & \textbf{42.33}
        & \textbf{20.03}
        & \textbf{12.52}
        & \textbf{29.40}
        & \textbf{42.93}
        & \textbf{13.80}
        
        & \textbf{50.33}
        & \textbf{20.65}
        & \textbf{31.04}
        & \textbf{18.81}
        \\

        \bottomrule
    \end{tabular}
\vskip-3ex
\end{table*}

\subsection{Quantitative Comparison.}
\label{sec:quantitative}

\noindent\textbf{Synthetic-Scene Evaluation on Pan-Replica.}
As shown in Table~\ref{tab:panreplica_main}, PanOVOcc achieves the best overall performance on Pan-Replica, reaching an occupancy IoU of $75.63$ and a semantic mIoU of $36.99$. 
Compared with the strongest training-free baseline FreeOcc~\cite{jiang2026freeocc}, PanOVOcc improves occupancy IoU by $20.03$ points and mIoU by $7.06$ points.
The improvement is particularly pronounced for large structural categories, indicating that native panoramic observations provide effective spatial evidence for constructing complete scene-level geometry. 
Nevertheless, performance on small or visually ambiguous categories remains uneven, with chair and bed showing lower accuracy than several baselines. Overall, these results demonstrate that PanOVOcc provides a strong training-free solution for persistent scene-level occupancy mapping, with substantial gains in both geometric reconstruction and semantic occupancy prediction.

\vspace{3pt}\noindent\textbf{Real-World Evaluation on Pan-Holo360D.}
Table~\ref{tab:panholo360d_main} reports quantitative results on the real-world Pan-Holo360D benchmark. 
PanOVOcc achieves an occupancy IoU of $66.27$ and semantic mIoU of $34.72$ under the eight-category evaluation protocol, outperforming both supervised and training-free baselines. 
Compared with FreeOcc~\cite{jiang2026freeocc}, our method improves occupancy IoU by $45.61$ points and semantic mIoU by $24.71$ points, demonstrating robust scene-level mapping under real-world panoramic observations. 
Consistent with Pan-Replica, the gains are particularly pronounced for large structural categories, highlighting the advantage of panoramic observations in recovering dominant scene geometry.
Moreover, PanOVOcc achieves high accuracy on several additional open-vocabulary categories without task-specific 3D training.
Overall, these results further demonstrate its effectiveness and generalizability for persistent open-vocabulary occupancy mapping in real-world indoor environments.

\vspace{3pt}\noindent\textbf{Spatial Coverage of Panoramic Observations.}
To characterize the spatial evidence provided by panoramic observations independently of the mapping algorithm, we conduct an observation-level coverage analysis under identical benchmark trajectories.
Let $\mathcal{V}_{i}^{s}$ denote the set of ground-truth occupied voxels directly observed at camera location $i$ under observation setting $s$.
We define the occupied coverage as
\setlength{\abovedisplayskip}{5pt}
\setlength{\belowdisplayskip}{5pt}
\begin{equation}
    \mathrm{Cov}(s)
    =
    \frac{
        \left|
        \bigcup_{i=1}^{N}
        \mathcal{V}_{i}^{s}
        \right|
    }{
        \left|
        \mathcal{V}_{\mathrm{occ}}^{\mathrm{GT}}
        \right|
    },
\end{equation}
where $\mathcal{V}_{\mathrm{occ}}^{\mathrm{GT}}$ denotes the set of ground-truth occupied voxels.

\begin{figure*}[!t]
    \centering

    \begin{minipage}[t]{0.64\textwidth}
        \vspace{0pt}
        \centering

        {\small\bfseries
            (a) Matched-progress mapping
        }
        \par\vspace{4pt}

        \footnotesize
        \setlength{\tabcolsep}{2pt}
        \renewcommand{\arraystretch}{1.0}

        \renewcommand{\tabularxcolumn}[1]{m{#1}}

        \begin{tabularx}{\linewidth}{
            @{}
            >{\centering\arraybackslash}m{0.4cm}
            @{\hspace{8pt}}
            *{3}{>{\centering\arraybackslash}X}
            @{}
        }

            \rotatebox[origin=c]{90}{\textbf{Perspective}}
            &
            \includegraphics[width=\linewidth]
                {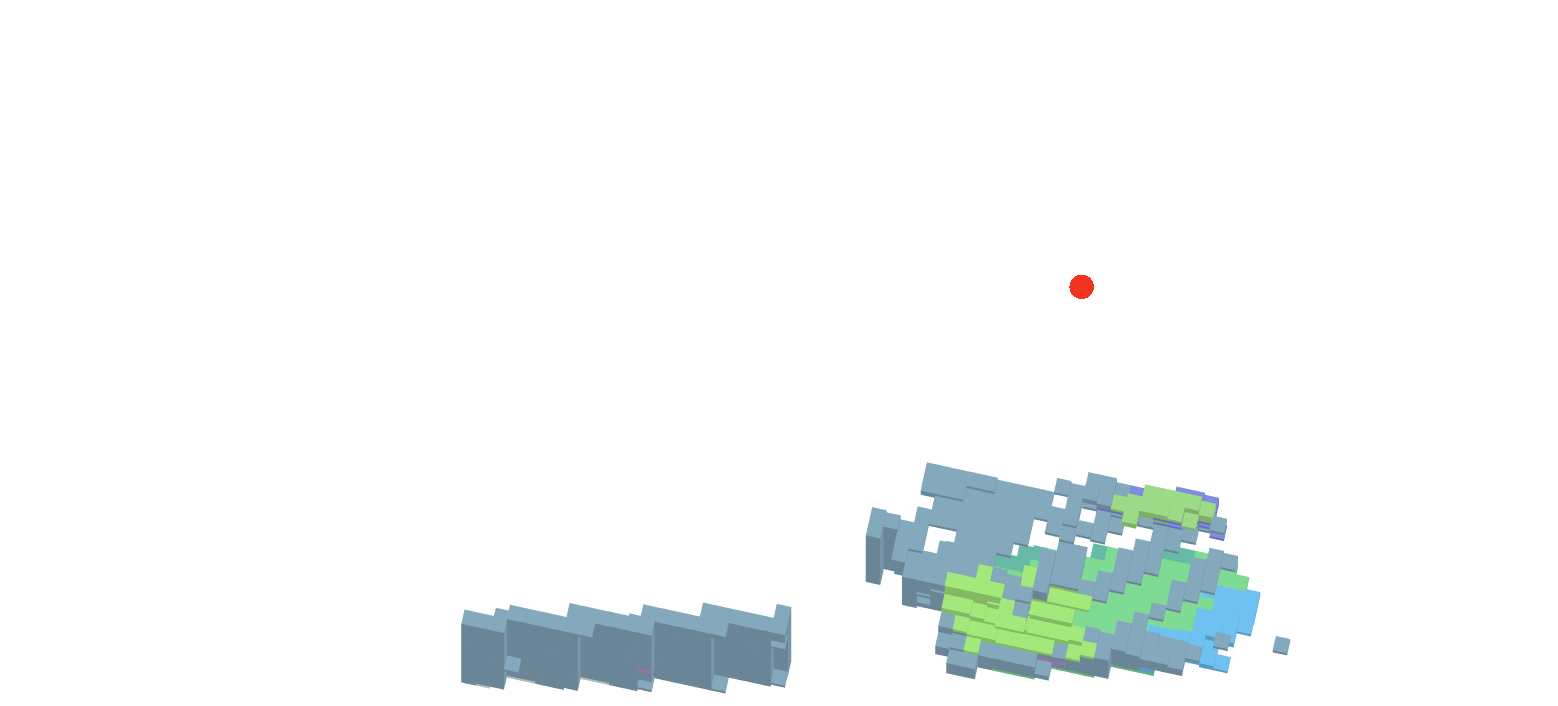}
            &
            \includegraphics[width=\linewidth]
                {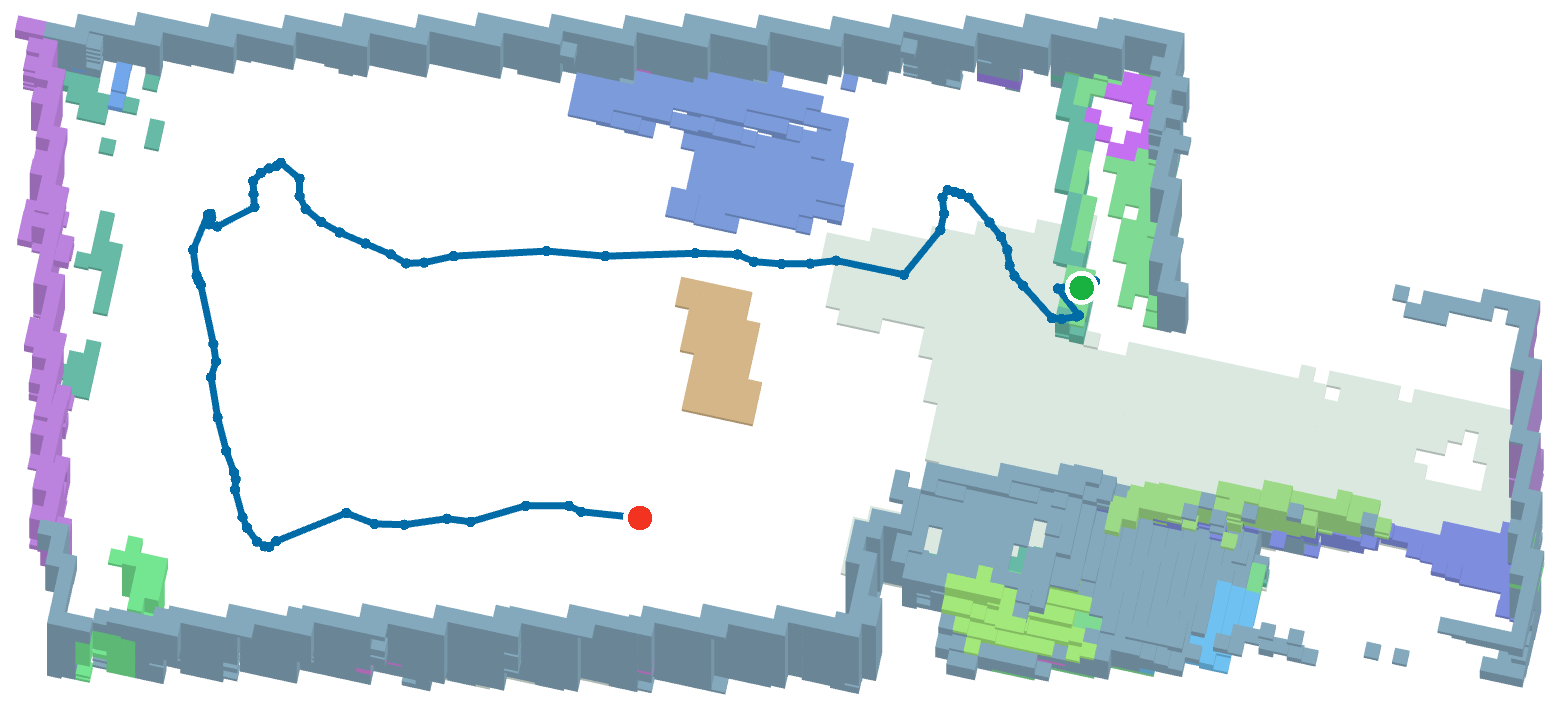}
            &
            \includegraphics[width=\linewidth]
                {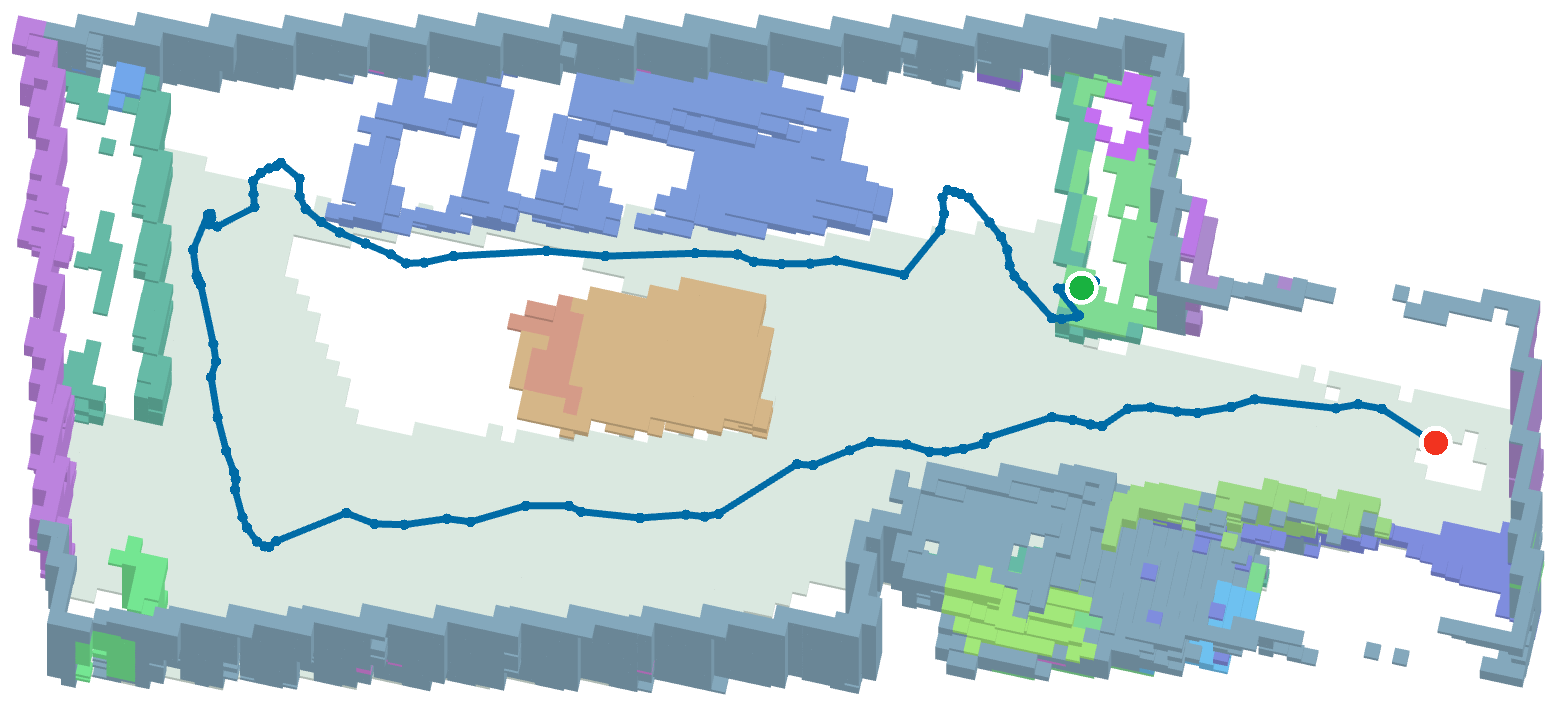}
            \\[-1pt]

            &
            8.24\% &
            53.22\% &
            67.84\%
            \\[5pt]

            \rotatebox[origin=c]{90}{\textbf{Panoramic}}
            &
            \includegraphics[width=\linewidth]
                {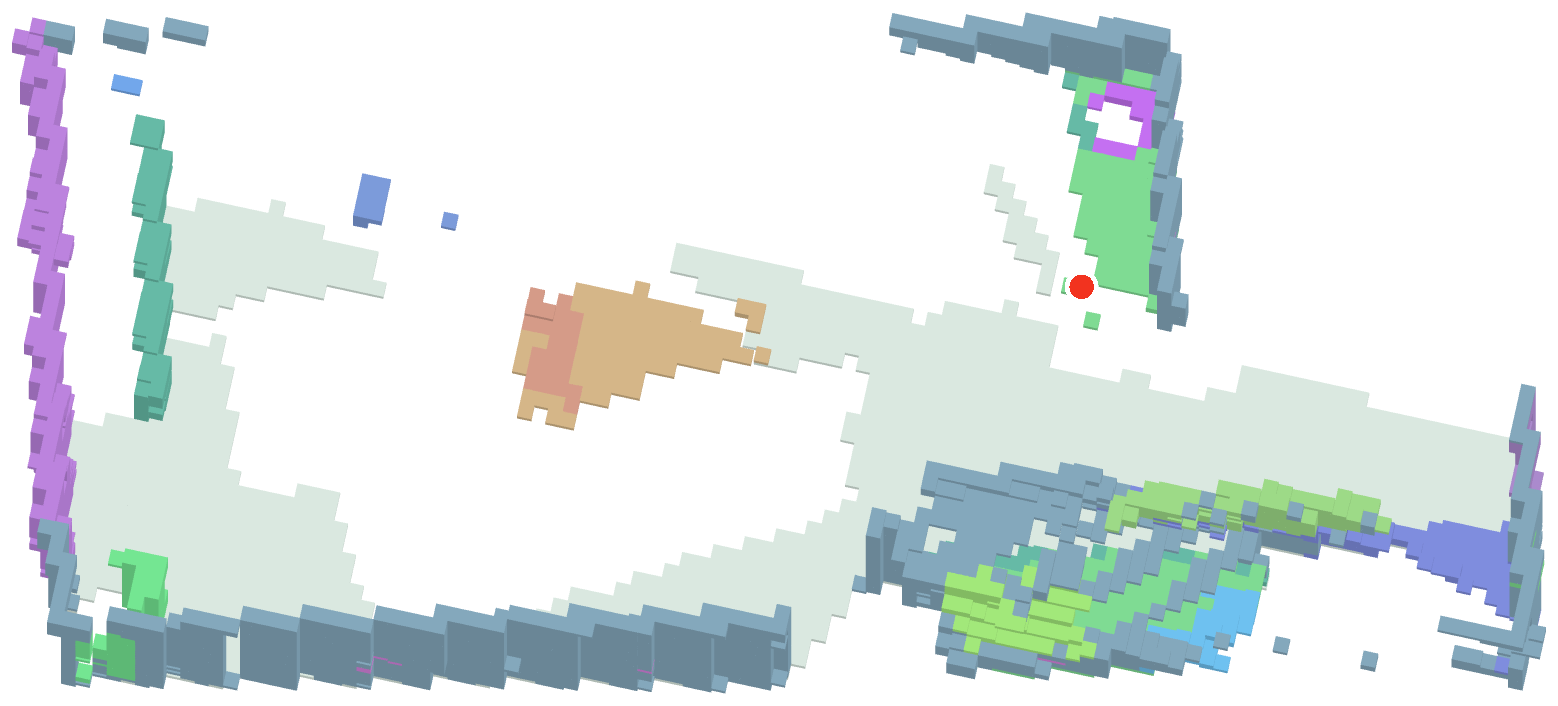}
            &
            \includegraphics[width=\linewidth]
                {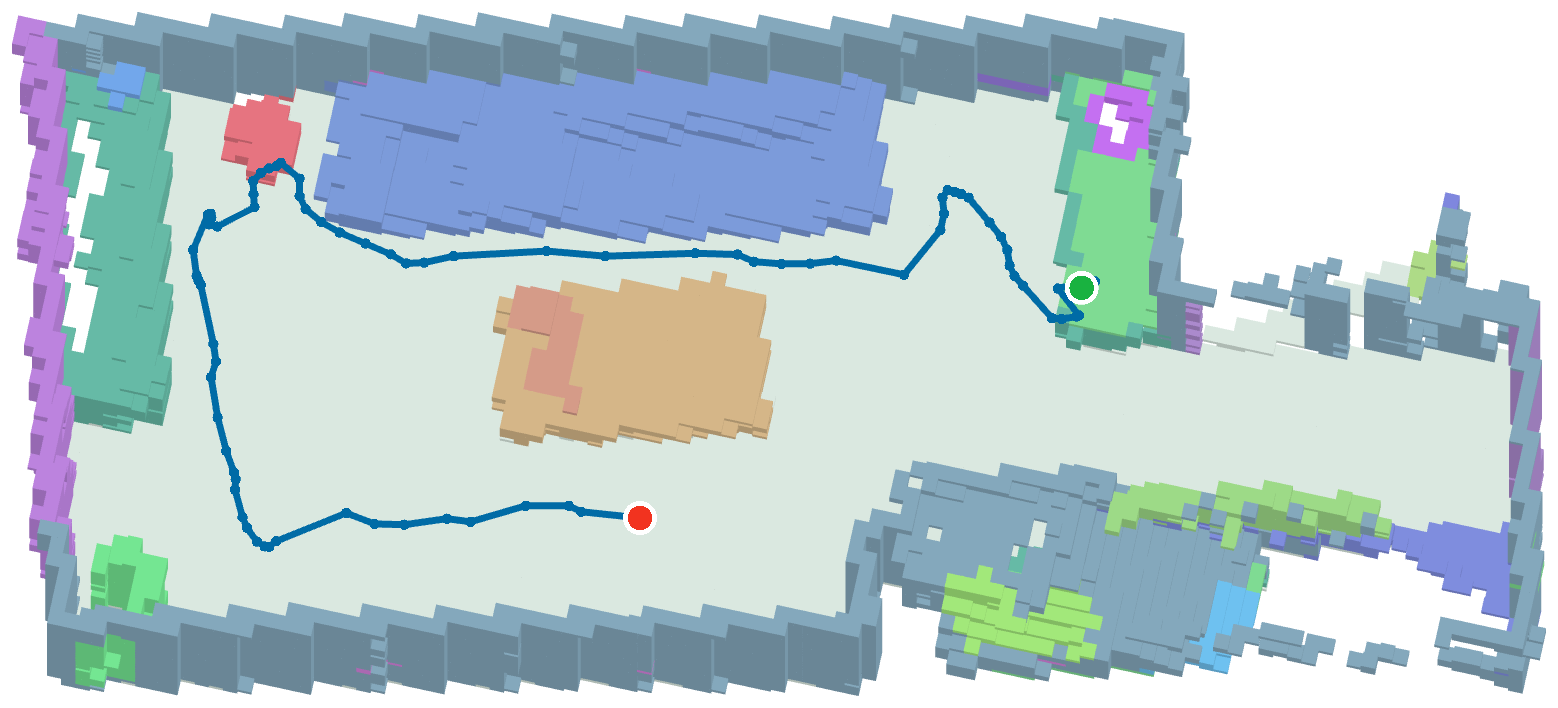}
            &
            \includegraphics[width=\linewidth]
                {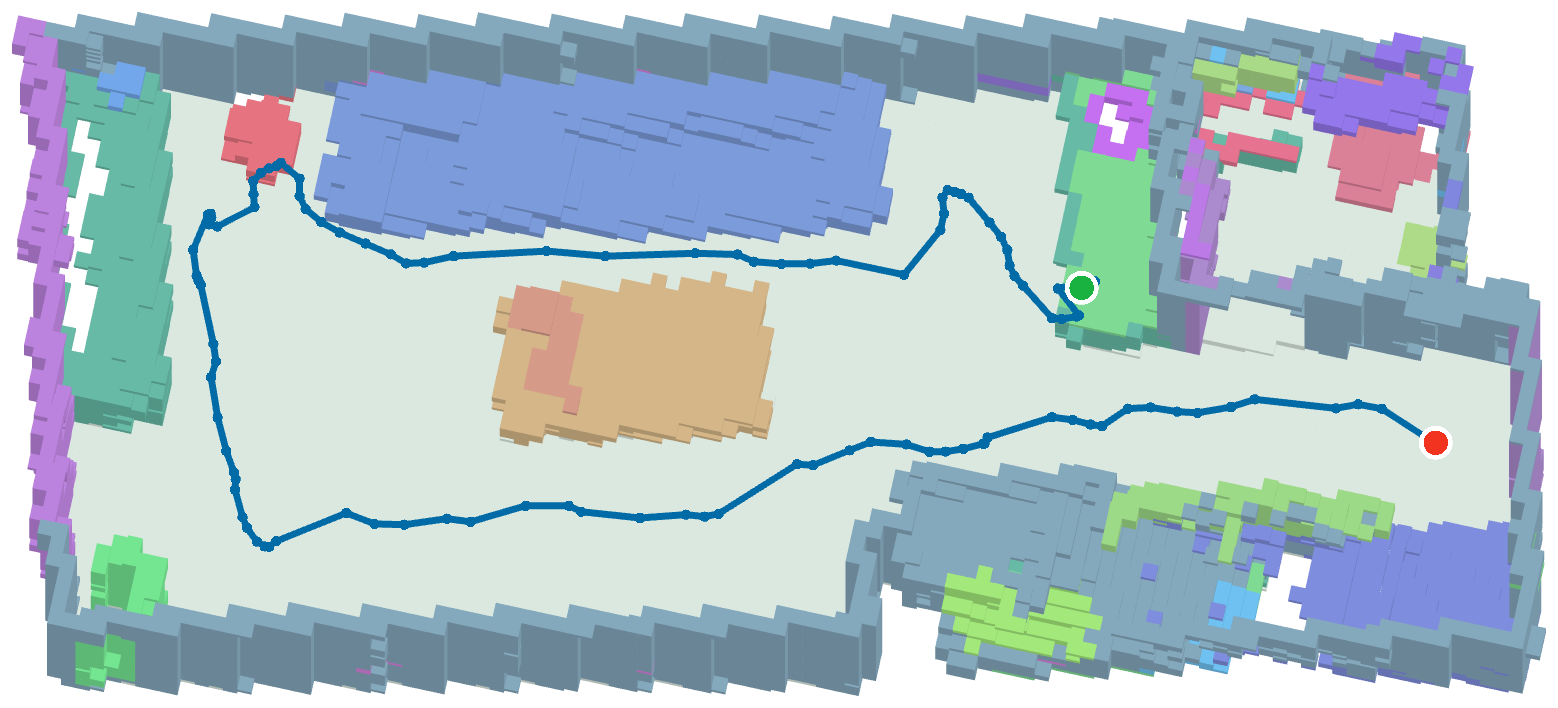}
            \\[-1pt]

            &
            45.15\% &
            81.58\% &
            94.29\%
            \\[4pt]

            &
            $N=1$ &
            $N=80$ &
            $N=110$

        \end{tabularx}
    \end{minipage}
    \hfill
    \begin{minipage}[t]{0.34\textwidth}
        \vspace{0pt}
        \centering

        {\small\bfseries
            (b) Coverage over the sequence
        }
        \par\vspace{4pt}

        \includegraphics[width=5.4cm]
            {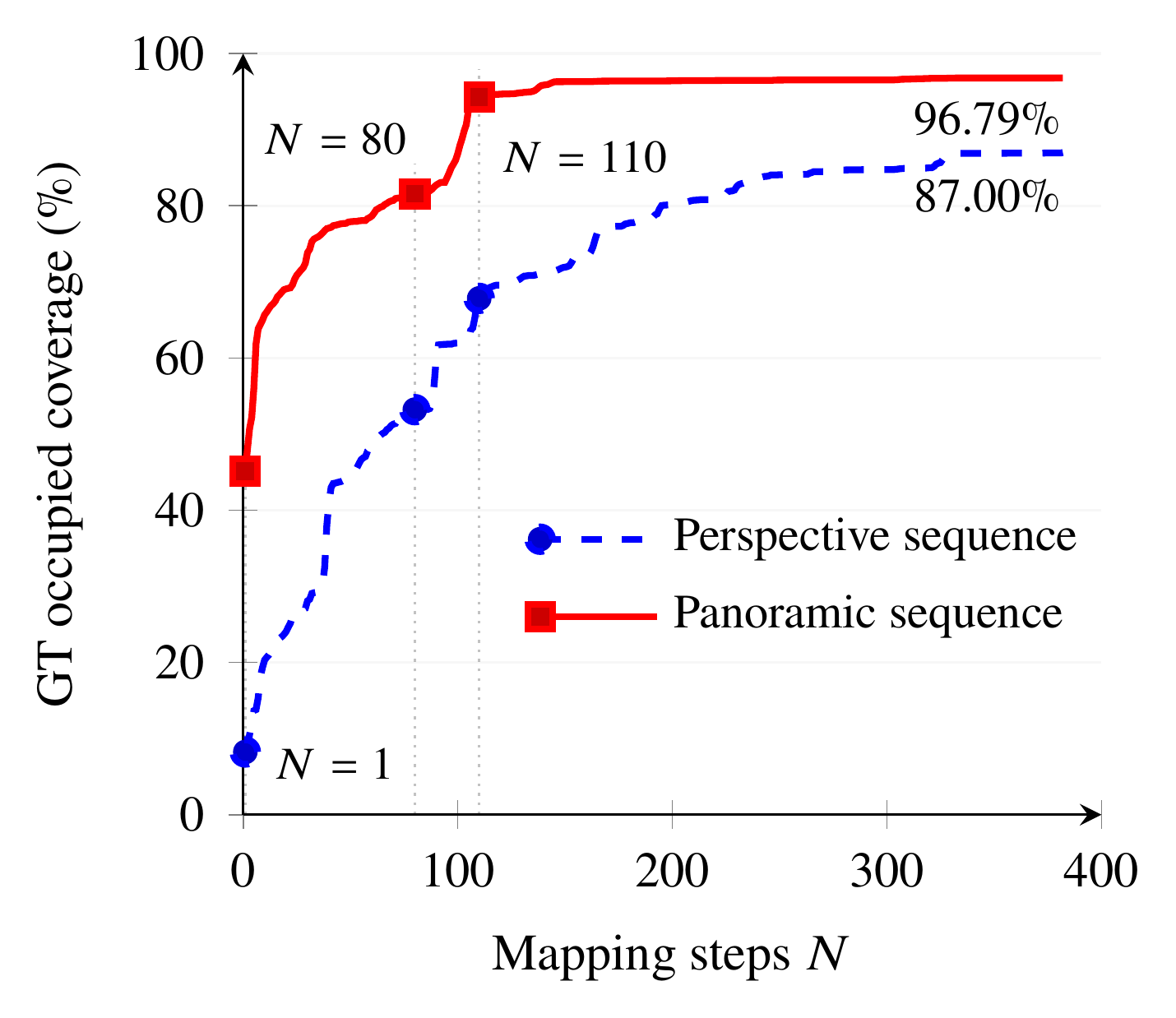}
    \end{minipage}

    \vskip-1ex
    \caption{
    \textbf{Observation-level spatial coverage.}
    Perspective and panoramic observations are compared along the same trajectory circling the room in \textit{Indoor\_015}.
    (a) Cumulative coverage visualizations at matched observation steps $N=1,80,110$.
    (b) Cumulative coverage over the complete sequence. Vertical dotted lines and markers correspond to the snapshots in (a).
    }
        
    \label{fig:observation_coverage}
    \vskip-3ex
\end{figure*}

Fig.~\ref{fig:observation_coverage} jointly presents matched-step coverage visualizations and cumulative coverage curves.
Results show that panoramic observations accumulate spatial evidence substantially faster than perspective observations.
At the first camera location, panoramic coverage reaches $45.15\%$, compared with $8.24\%$ for perspective observations, illustrating the additional scene geometry exposed by the wider field of view.
As the camera moves around the scene, this advantage translates into more complete cumulative coverage.
By $N{=}110$, the panoramic sequence observes $94.29\%$ of the ground-truth occupied voxels, whereas the perspective sequence covers only $67.84\%$, leaving nearly one third unobserved.
The corresponding visualizations also illustrate this contrast.
Panoramic observations provide near-complete coverage of the occupied scene, while substantial regions remain unobserved under the perspective setting despite following the same trajectory.
These results highlight an observation-level advantage of panoramic view, showing that the wider field of view provides more comprehensive spatial evidence.

\subsection{Ablation Studies}
\label{sec:ablation}

We analyze both the internal components of PanOVOcc and its dependence on upstream perception modules.
%

\vspace{3pt}\noindent\textbf{Component Ablation.}
We first evaluate the contribution of the three main components.
For ablation studies, we reuse the same SLAM trajectory while introducing predefined pose perturbations and corrections to evaluate the effect of each component under a consistent protocol. Since this setting differs from the default online mapping pipeline in the pose-update schedule and map-maintenance procedure, absolute scores are not directly compared with those in the main experiment; all component gains are assessed only within the ablation setting.
Table~\ref{tab:component_ablation} summarizes the contribution of each component. Removing PSVF leads to the largest performance drop, reducing IoU and mIoU to $53.02$ and $23.51$, respectively. GSC improves semantic performance from $31.51$ to $32.09$ mIoU without affecting occupancy IoU, while RPU further improves both occupancy and semantic results. 
These results confirm the complementary contributions of PSVF, GSC, and RPU to geometric fusion, semantic consolidation, and persistent map maintenance.

\begin{table}[!t]
    \centering
    \caption{
        Component ablation of PanOVOcc on Pan-Replica.
        PSVF, GSC, RPU denote Panoramic Signed Voxel Fusion, Geometry-Aware Semantic Consolidation, and Reversible Persistent Update.
    }
    \label{tab:component_ablation}
    \vskip -1ex
    \footnotesize
    \setlength{\tabcolsep}{5.0pt}
    \renewcommand{\arraystretch}{1.08}
    \begin{tabular}{@{}lccccc@{}}
        \toprule
        \textbf{Variant}
        & \textbf{PSVF}
        & \textbf{GSC}
        & \textbf{RPU}
        & \textbf{IoU $\uparrow$}
        & \textbf{mIoU $\uparrow$}
        \\
        \midrule
        w/o PSVF
        & \xmarkblack 
        & \cmarkblack 
        & \cmarkblack
        & 53.02 & 23.51
        \\
        w/o GSC
        & \cmarkblack 
        & \xmarkblack 
        & \cmarkblack
        & \textbf{76.75} & 31.51
        \\
        w/o RPU
        & \cmarkblack 
        & \cmarkblack 
        & \xmarkblack
        & 69.69 & 30.54
        \\
        \midrule
        \textbf{Full model} 
        & \cmarkblack 
        & \cmarkblack 
        & \cmarkblack
        & \textbf{76.75} & \textbf{32.09} \\
        \bottomrule
    \end{tabular}
    \vskip -4ex
\end{table}

\vspace{3pt}\noindent\textbf{Effect of Upstream Perception Modules.}
We evaluate how the SLAM frontend and semantic predictor affect mapping performance by replacing each module separately. These ablations use controlled final-pose replay with fixed mapping frames and voxel coordinates. As this protocol differs from that of the main comparison, baseline scores may differ, and module effects are evaluated only within each matched ablation group.
Table~\ref{tab:perception_ablation} (a) shows that ODGS-SLAM provides the highest geometric IoU among estimated-pose frontends on both benchmarks.
On Pan-Replica, it outperforms OpenVSLAM and its dense fork by $17.63$ and $15.59$ in IoU, respectively.
On Pan-Holo360D, the corresponding gains are $7.29$ and $23.97$ points.
These results support the importance of pose accuracy for aligning observations during geometric fusion.
In contrast, replacing the semantic predictor leaves geometric IoU unchanged in Table~\ref{tab:perception_ablation} (b),
indicating that frontend poses mainly govern geometric consistency, while the semantic predictor affects semantic occupancy quality.

\subsection{Qualitative Analysis}
\label{sec:qualitative}

\begin{figure*}[t]
    \centering
    \setlength{\tabcolsep}{2pt}

    \begin{tabularx}{\textwidth}{
        @{}
        >{\centering\arraybackslash}X
        >{\centering\arraybackslash}X
        >{\centering\arraybackslash}X
        >{\centering\arraybackslash}X
        >{\centering\arraybackslash}X
        >{\centering\arraybackslash}X
        >{\centering\arraybackslash}X
        @{}
    }
        \textbf{EmbodiedOcc} &
        \textbf{EmbodiedOcc++} &
        \textbf{VEOcc} &
        \textbf{GPOcc} &
        \textbf{FreeOcc} &
        \textbf{Ours} &
        \textbf{GT}
        \\[2pt]

        \includegraphics[height=2cm]{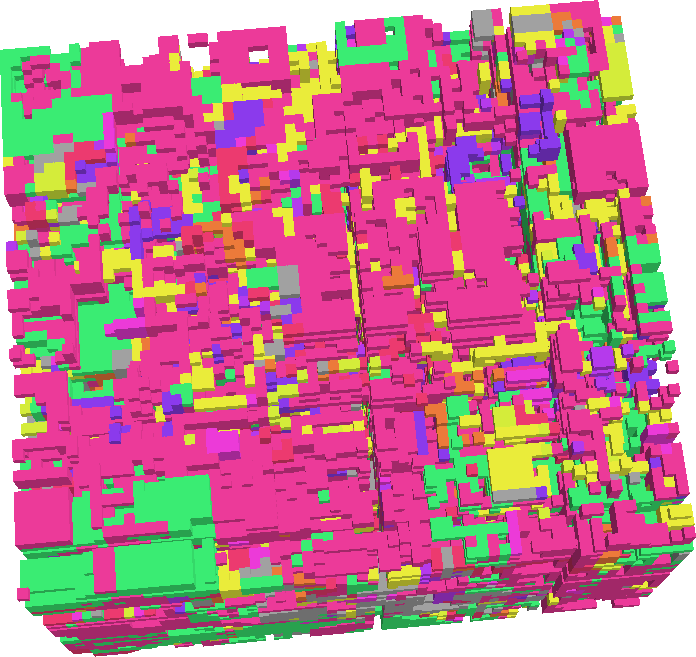} &
        \includegraphics[height=2cm]{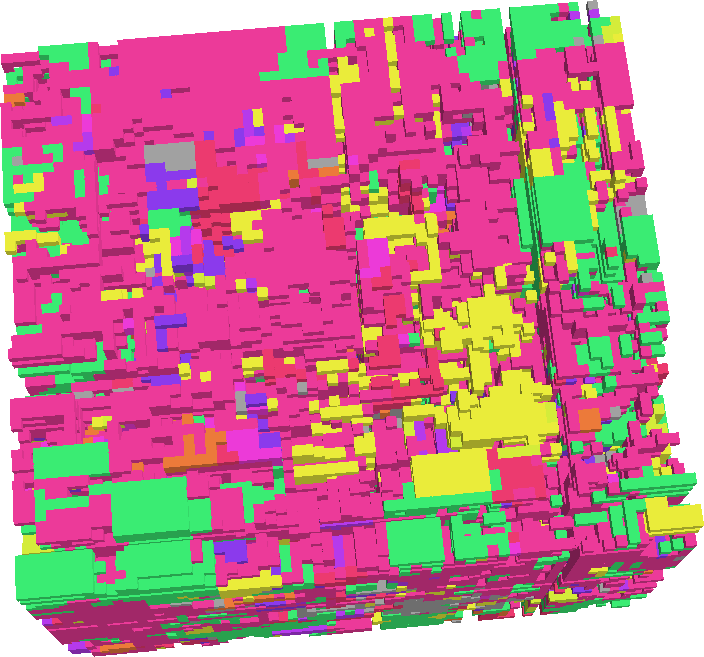} &
        \includegraphics[height=2cm]{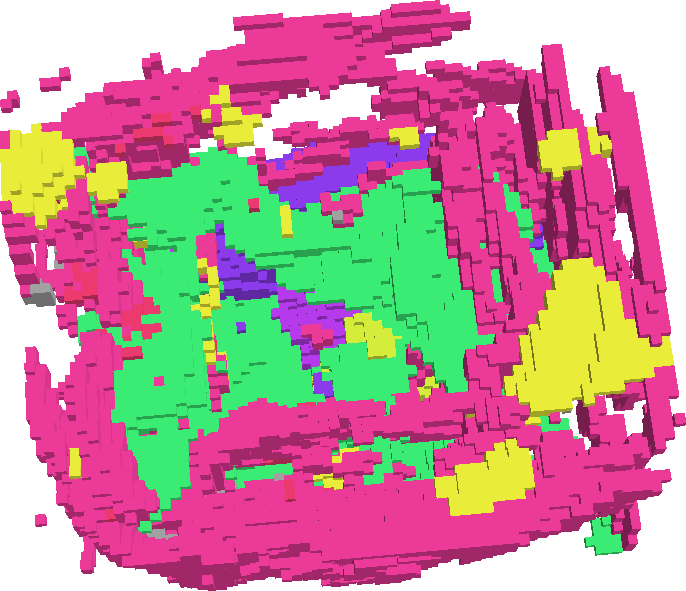} &
        \includegraphics[height=2cm]{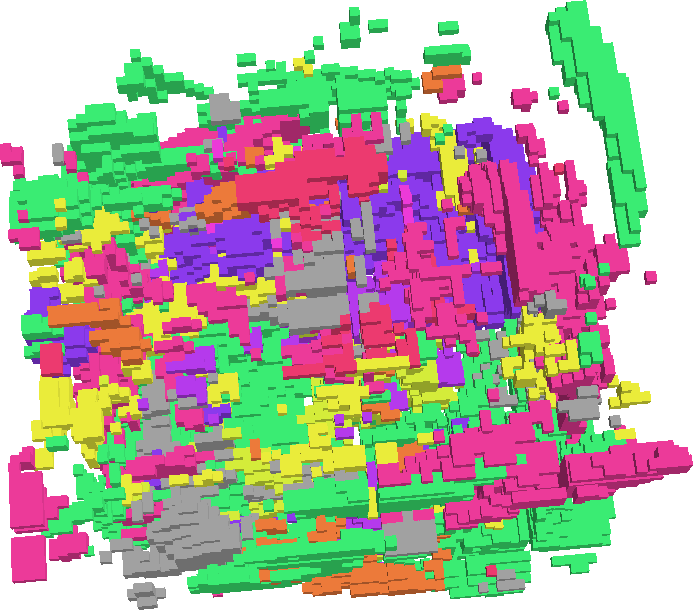} &
        \includegraphics[height=2cm]{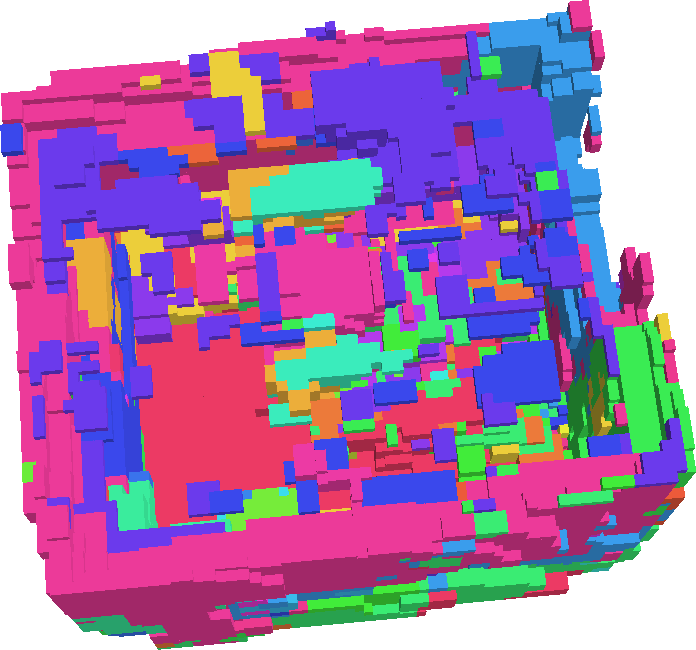} &
        \includegraphics[height=2cm]{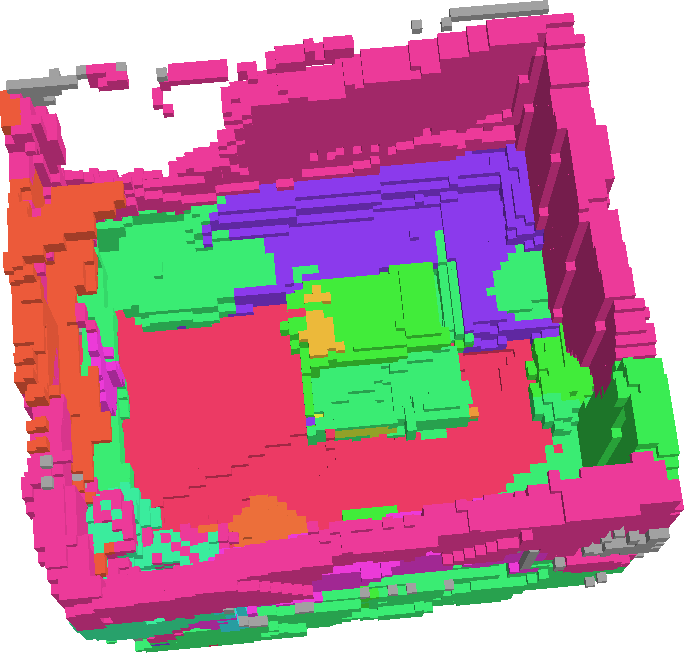} &
        \includegraphics[height=2cm]{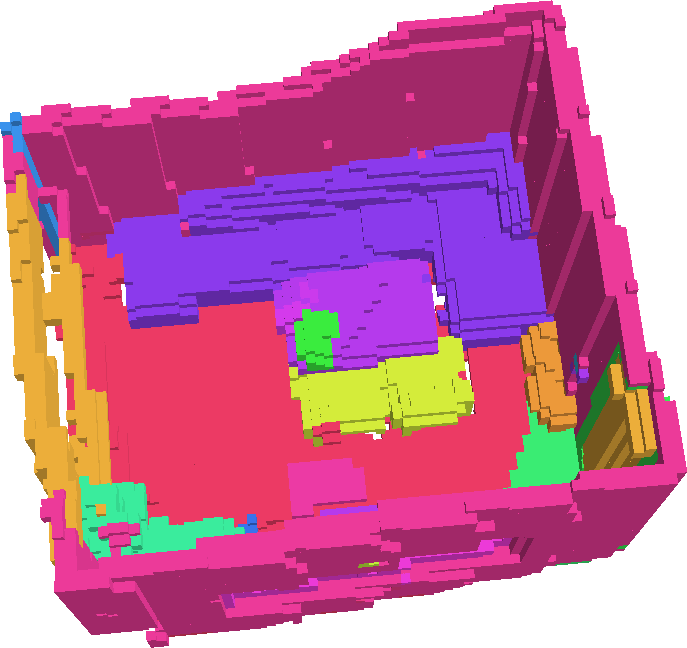}
        \\[-2pt]

        \includegraphics[width=\linewidth]{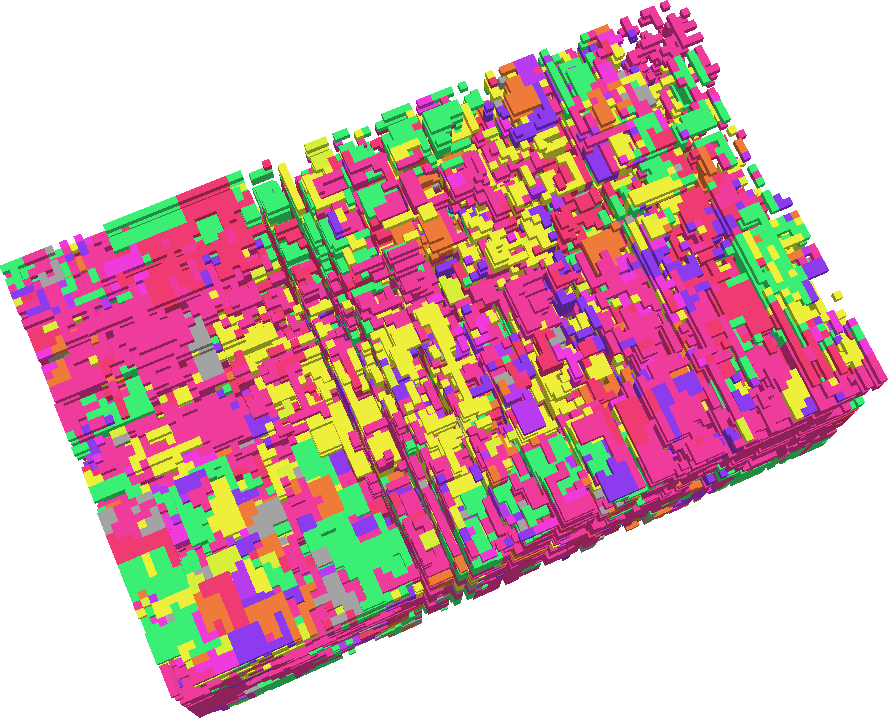} &
        \includegraphics[width=\linewidth]{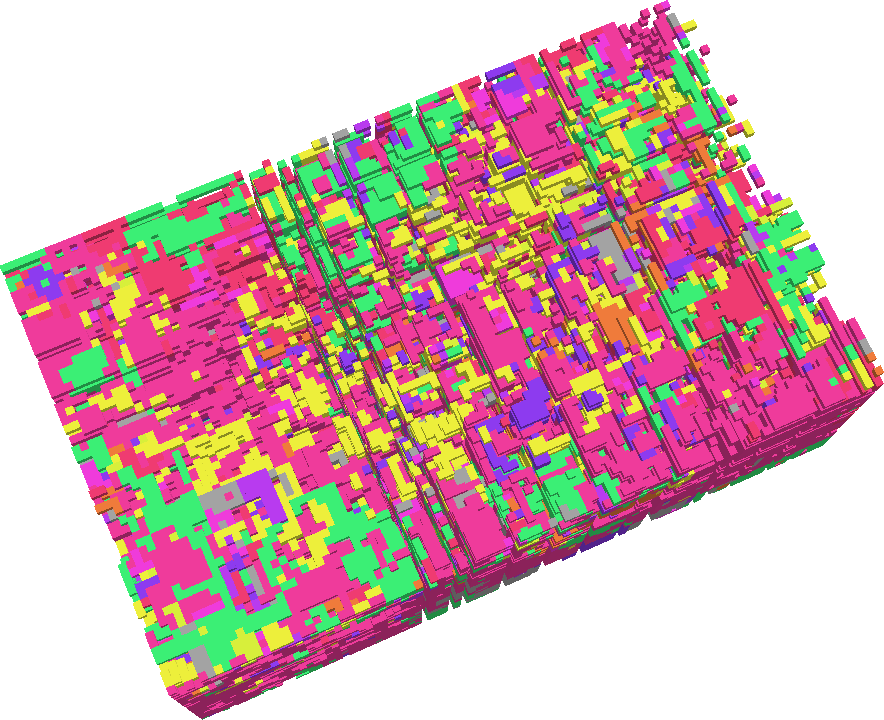} &
        \includegraphics[width=\linewidth]{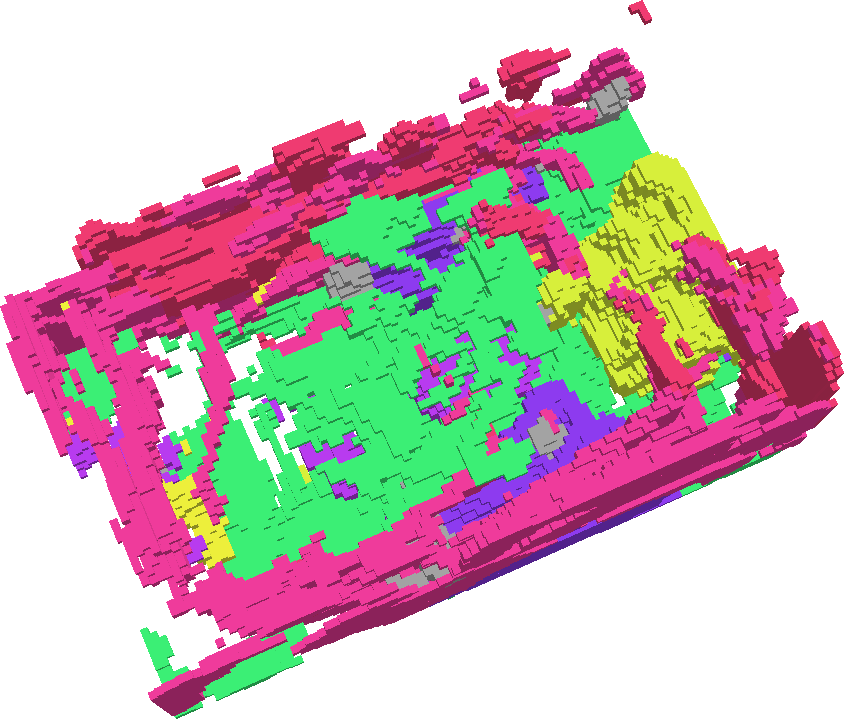} &
        \includegraphics[width=\linewidth]{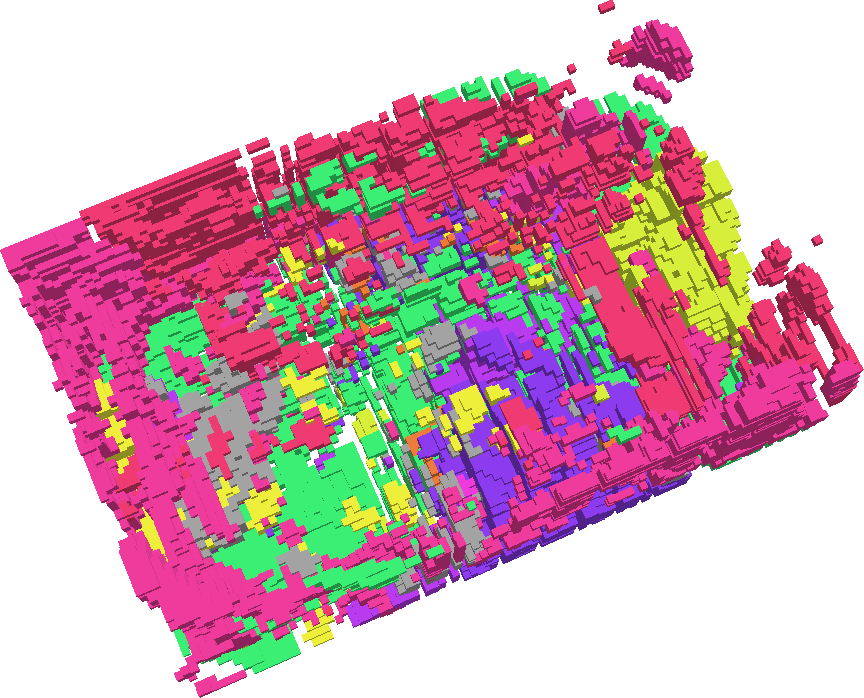} &
        \includegraphics[width=\linewidth]{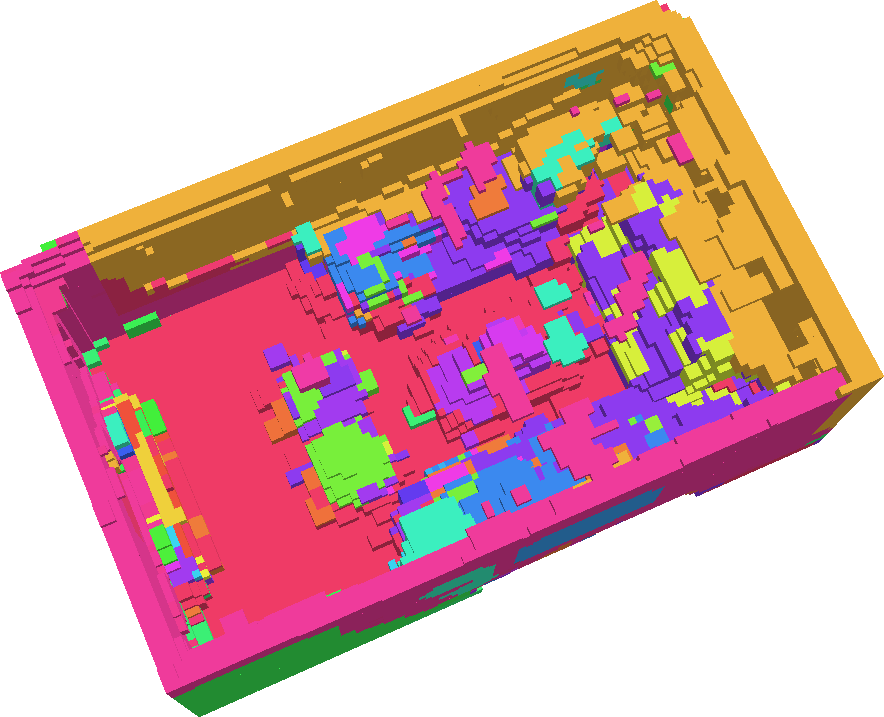} &
        \includegraphics[width=\linewidth]{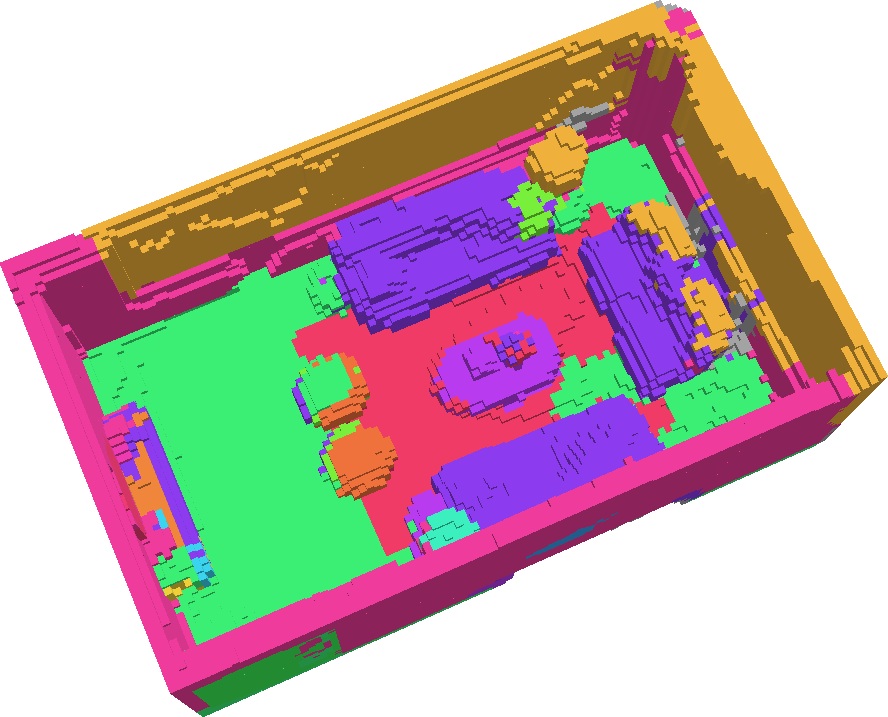} &
        \includegraphics[width=\linewidth]{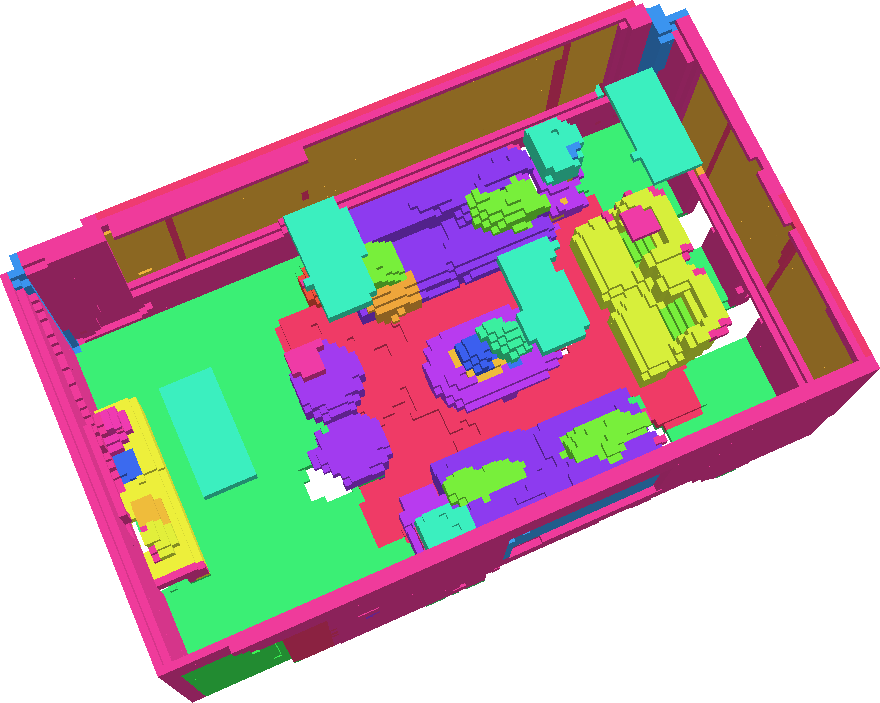}
        \\[0pt]

        \multicolumn{7}{c}{\small\textbf{(a)}}
    \end{tabularx}

    \vspace{0pt}

    \vspace{1pt}

    \begin{tabularx}{\textwidth}{
        @{}
        >{\centering\arraybackslash}X
        >{\centering\arraybackslash}X
        >{\centering\arraybackslash}X
        >{\centering\arraybackslash}X
        >{\centering\arraybackslash}X
        >{\centering\arraybackslash}X
        @{}
    }
        \textbf{EmbodiedOcc++} &
        \textbf{VEOcc} &
        \textbf{GPOcc} &
        \textbf{FreeOcc} &
        \textbf{Ours} &
        \textbf{GT}
        \\[0pt]

        \includegraphics[height=1.7cm]{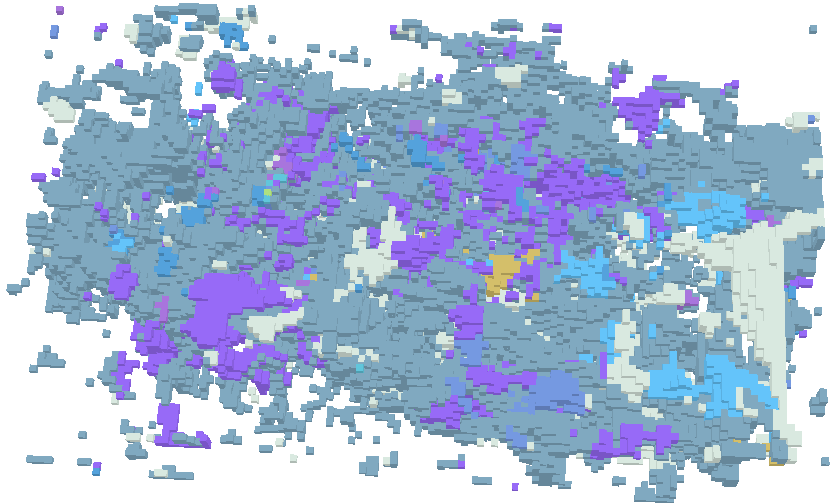} &
        \includegraphics[height=1.7cm]{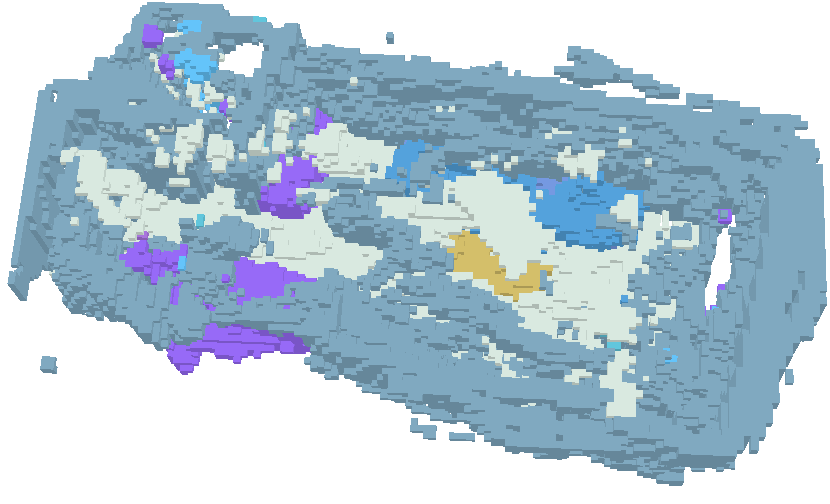} &
        \includegraphics[height=1.7cm]{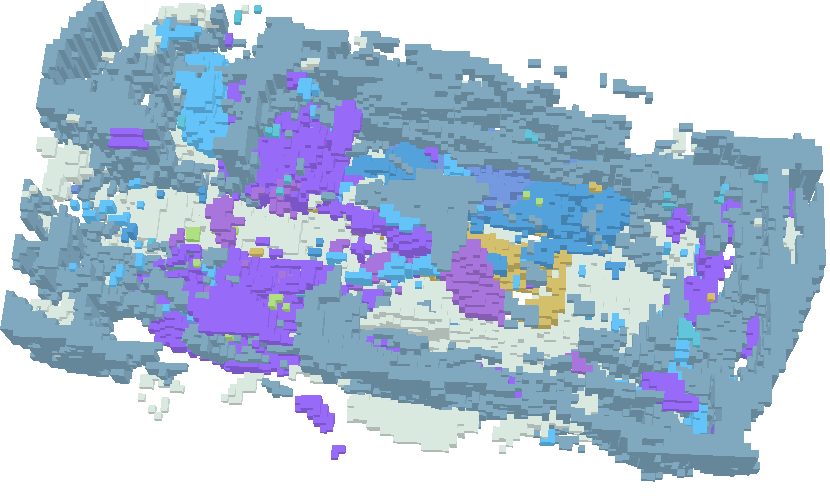} &
        \includegraphics[height=1.7cm]{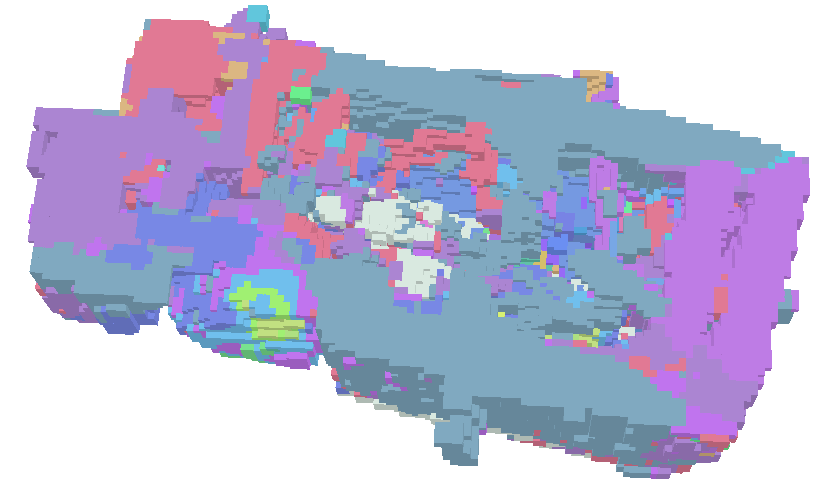} &
        \includegraphics[height=1.8cm]{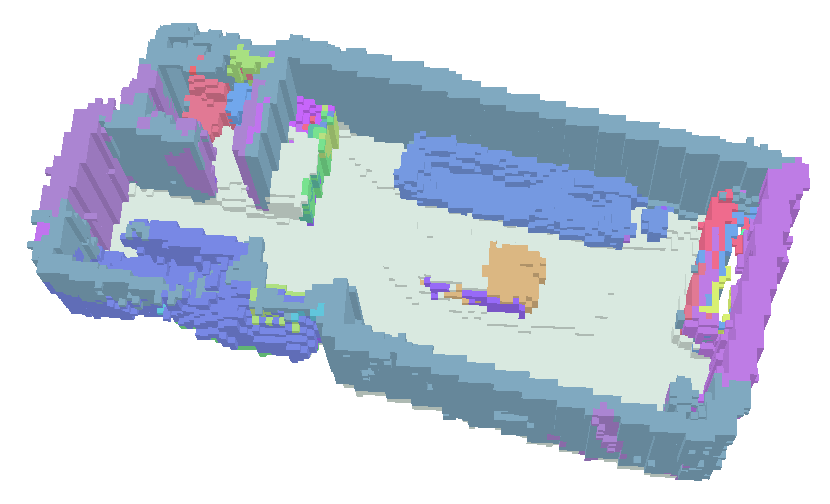} &
        \includegraphics[height=1.8cm]{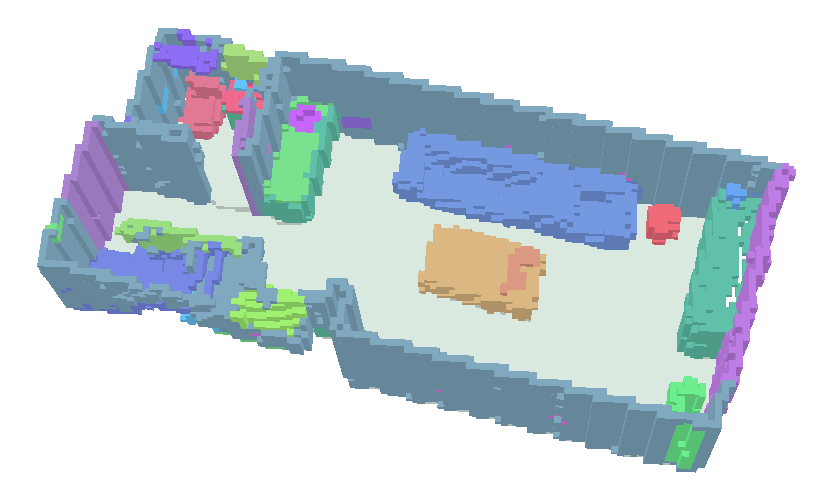}
        \\[-15pt]

        \includegraphics[height=2.3cm]{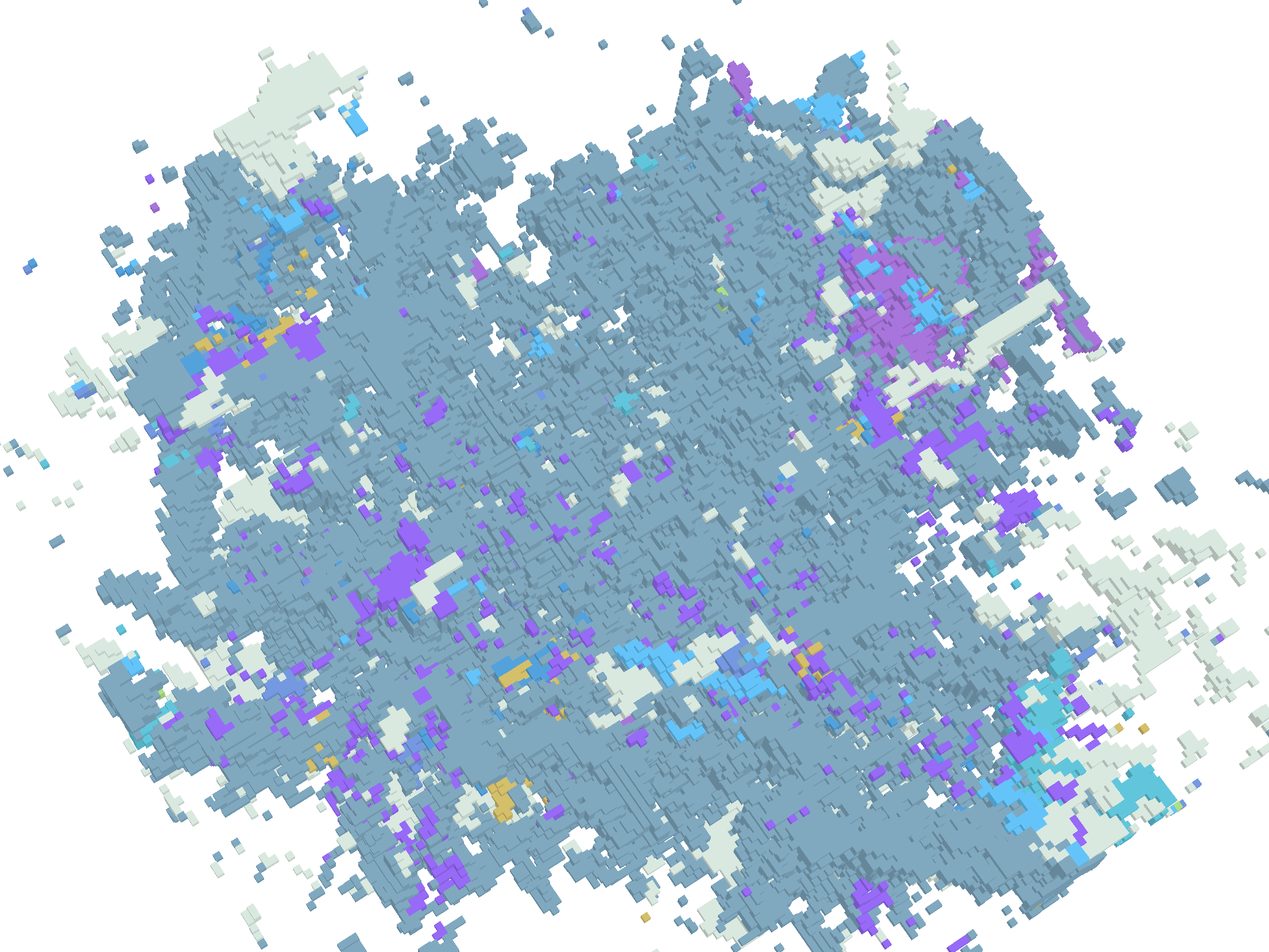} &
        \includegraphics[height=2.3cm]{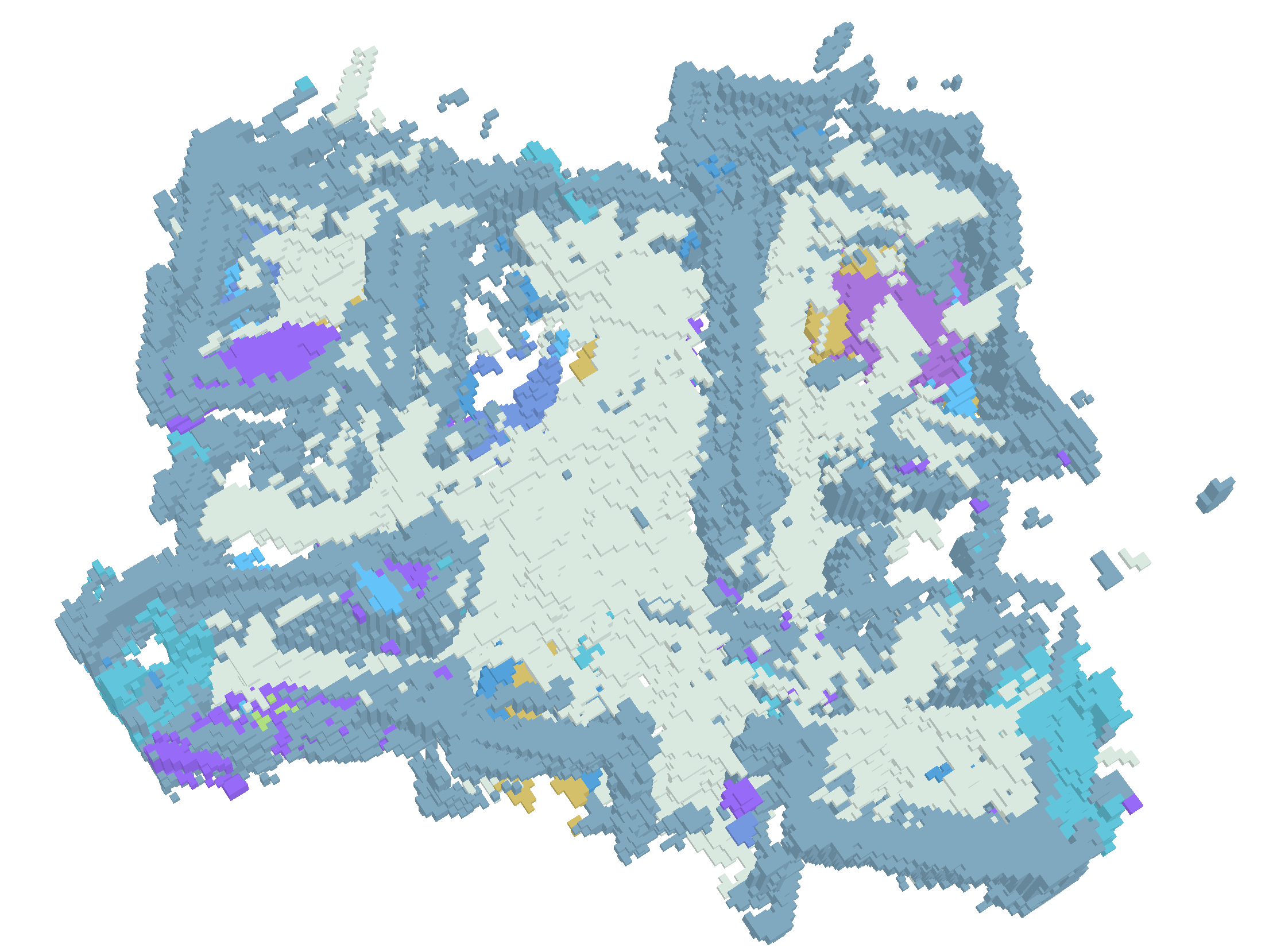} &
        \includegraphics[height=2.3cm]{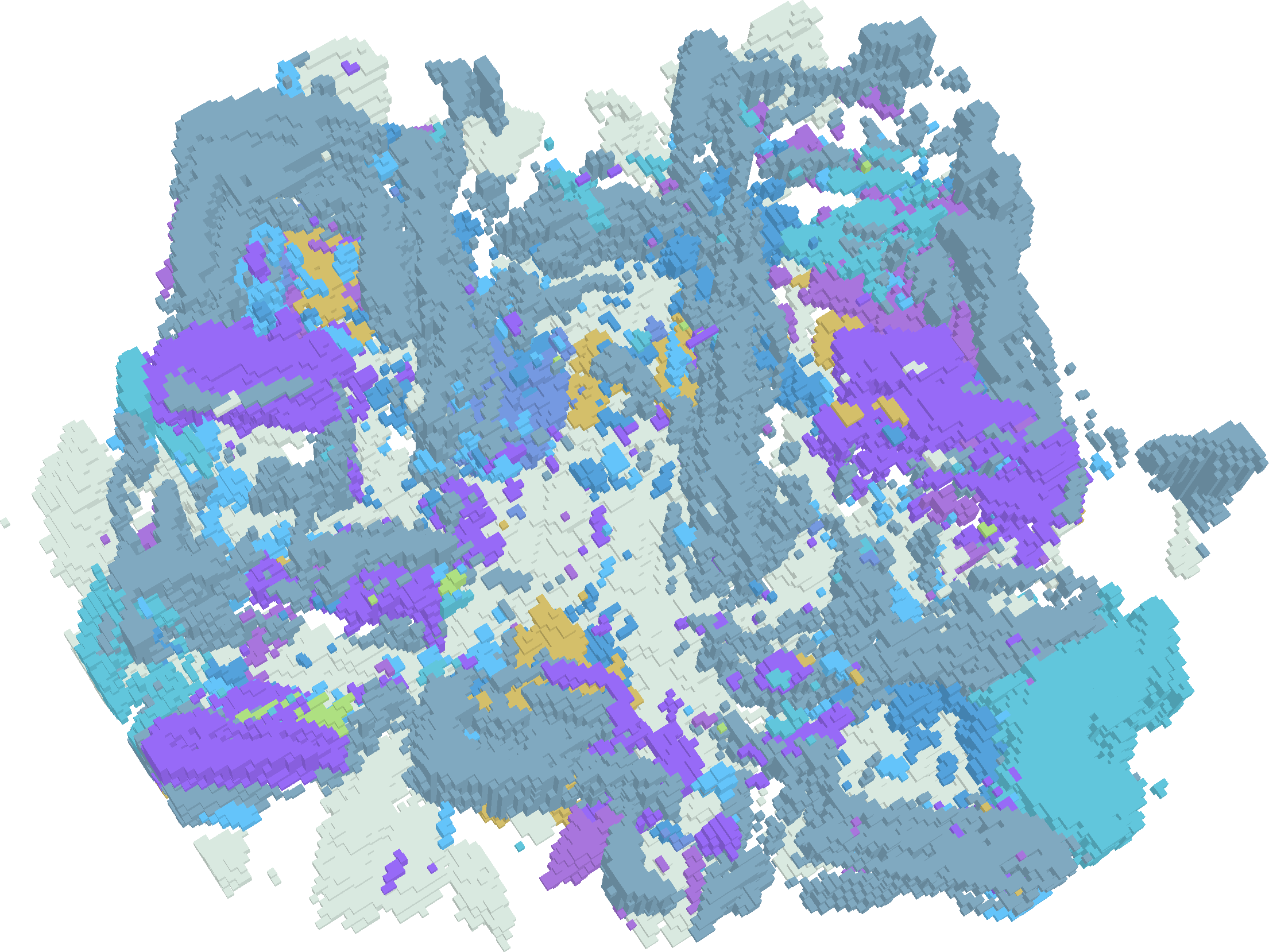} &
        \includegraphics[height=2.8cm]{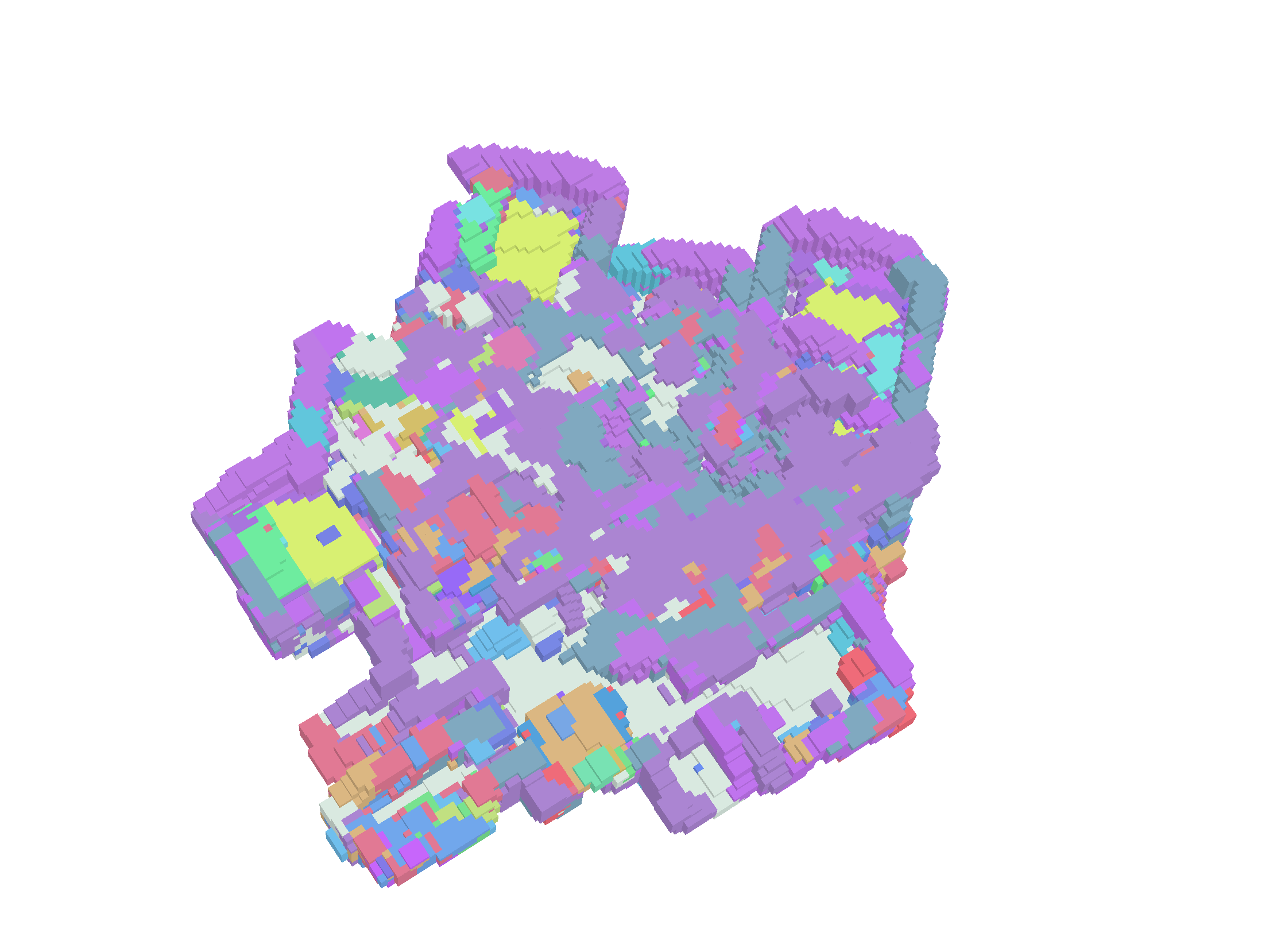} &
        \includegraphics[height=2.4cm]{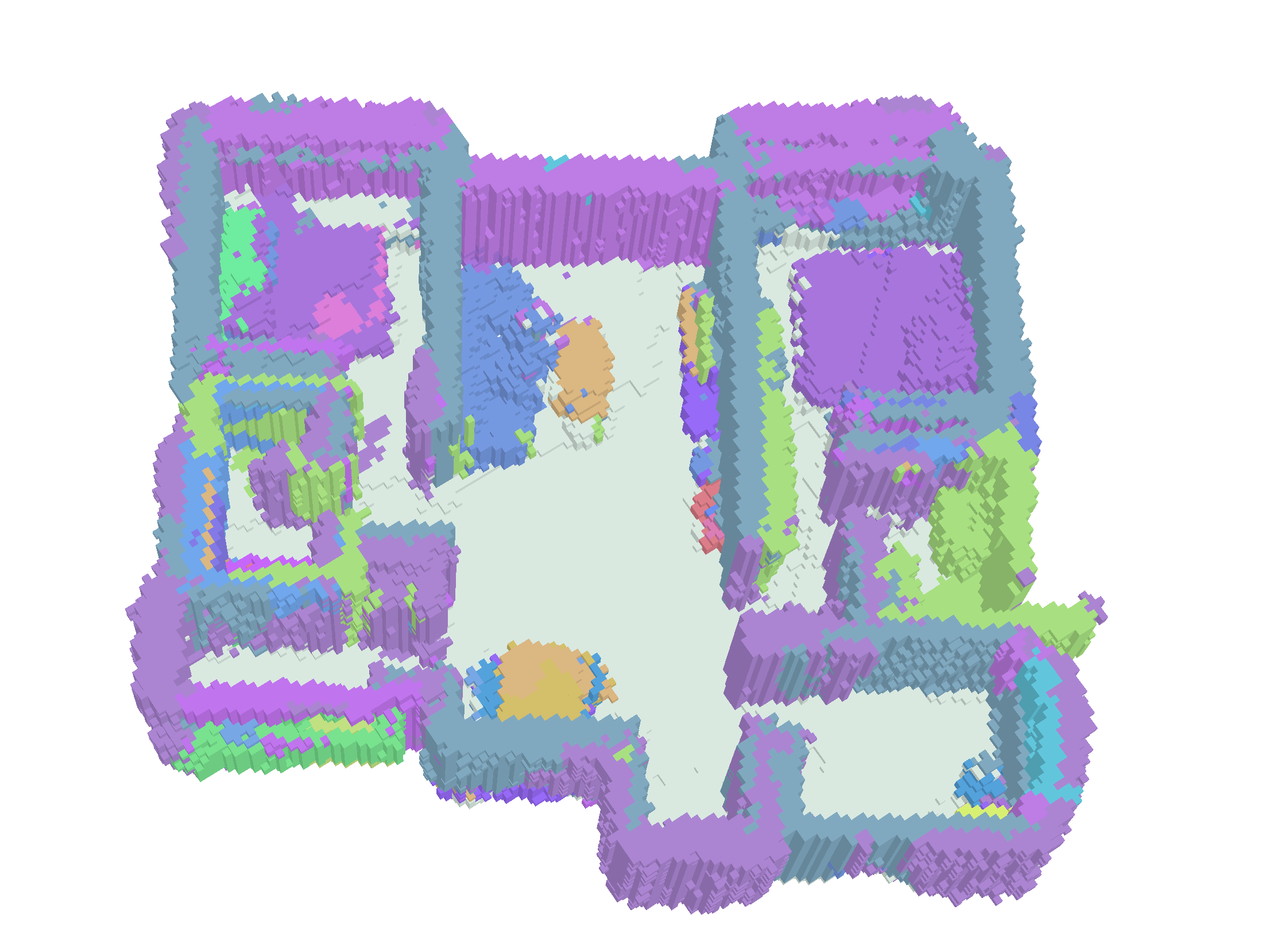} &
        \includegraphics[height=2.5cm]{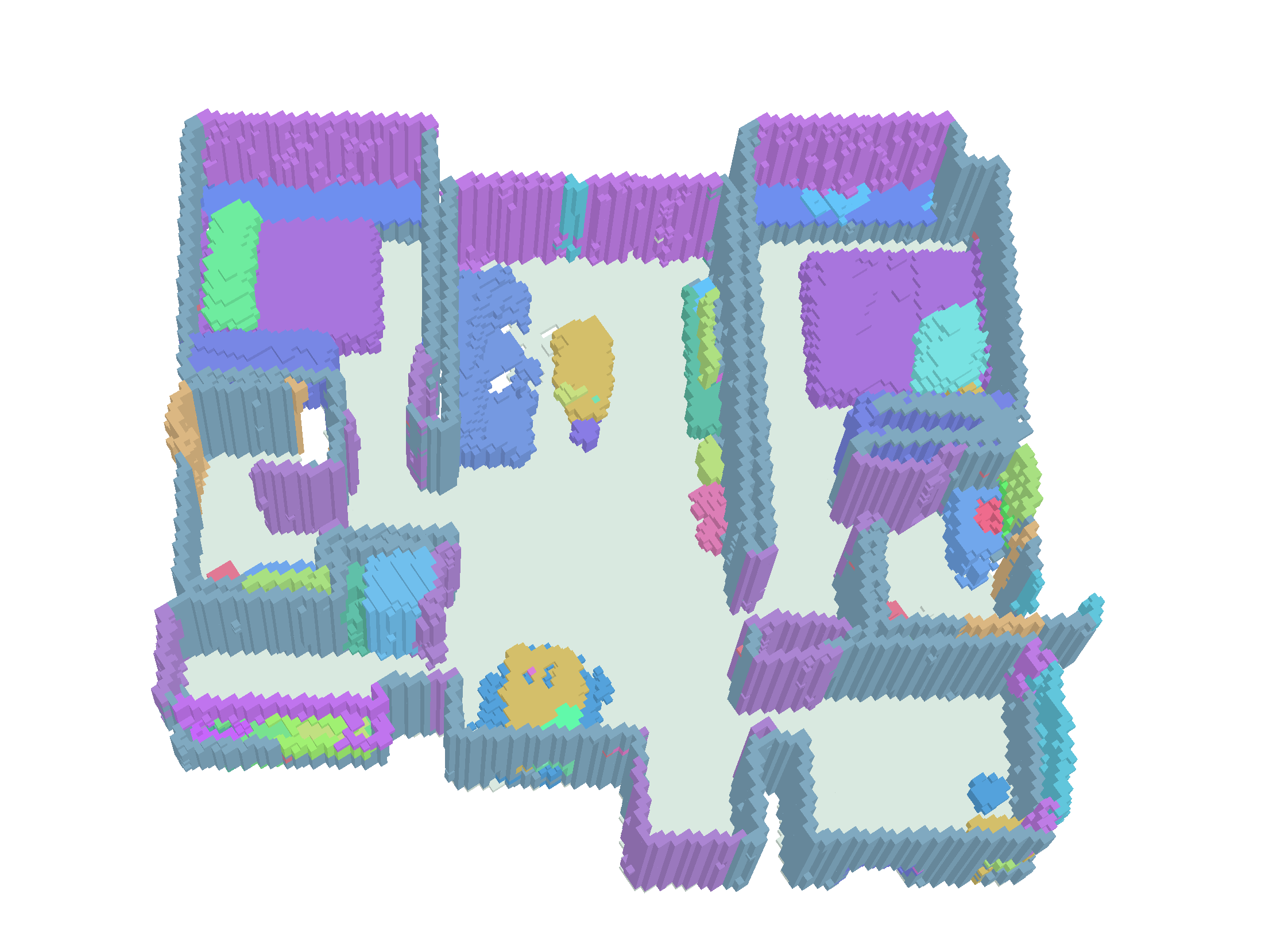}
        \\[-2pt]

        \multicolumn{6}{c}{\small\textbf{(b)}}
    \end{tabularx}
    \vskip-1ex
    \caption{
        \textbf{Qualitative comparison of semantic occupancy mapping results on (a) Pan-Replica and (b) Pan-Holo360D.}
        Compared with existing methods, \textit{e.g.}, EmbodiedOcc~\cite{wu2025embodiedocc}, EmbodiedOcc++~\cite{wang2025embodiedocc++}, VEOcc~\cite{wang2026veocc}, GPOcc~\cite{zhou2026gpocc}, and FreeOcc~\cite{jiang2026freeocc}, our approach more accurately reconstructs scene geometry and maintains strong performance in larger-scale environments. For clearer visualization, top-layer voxels are removed to reduce occlusion.
    }
    \label{fig:qualitative_results}
    \vskip -3ex
\end{figure*}

We provide qualitative comparisons on Pan-Replica and Pan-Holo360D in Fig.~\ref{fig:qualitative_results} to complement the quantitative evaluation.  
On Pan-Replica, PanOVOcc reconstructs clearer room boundaries and furniture layouts, whereas existing methods exhibit varying degrees of occupancy noise and geometric distortion.
This contrast becomes more pronounced in the larger real-world examples from Pan-Holo360D.
Specifically, our method better preserves the sofa and surrounding floor layout in the upper row of Fig.~\ref{fig:qualitative_results} (b), as well as room partitions around the central connecting area in the lower-row multi-room example, where several baselines produce fragmented or distorted layouts.
These examples suggest that preserving semantic targets and spatial context could support target localization and approach planning, while coherent room layouts could facilitate cross-room navigation.

\begin{table}[!t]
    \centering
    \caption{
        Effect of upstream perception modules.
        The default configuration uses ODGS-SLAM~\cite{spiss2026odgsslam} and OOOPS~\cite{zheng2024open}.
    }
    \label{tab:perception_ablation}
    \vskip-1ex

    \footnotesize
    \setlength{\tabcolsep}{3.0pt}
    \renewcommand{\arraystretch}{1.08}

    \begin{tabular}{@{}lcccc@{}}
        \toprule
        &
        \multicolumn{2}{c}{\textbf{Pan-Replica}}
        &
        \multicolumn{2}{c}{\textbf{Pan-Holo360D}}
        \\
        \cmidrule(lr){2-3}
        \cmidrule(lr){4-5}

        \textbf{Method}
        & \textbf{IoU $\uparrow$}
        & \textbf{mIoU $\uparrow$}
        & \textbf{IoU $\uparrow$}
        & \textbf{mIoU $\uparrow$}
        \\
        \midrule

        \multicolumn{5}{@{}l}{\textit{Default setting}} \\

        ours
        & \textbf{77.88}
        & 33.54
        & 65.64
        & \textbf{33.71}
        \\

        \midrule

        \multicolumn{5}{@{}l}{\textit{(a) SLAM variants}} \\
        
        OpenVSLAM~\cite{sumikura2019openvslam}
        & 60.25
        & 27.15
        & 58.35
        & 29.35
        \\
        OpenVSLAM dense~\cite{surmann2022vslamdense}
        & 62.29
        & 27.65
        & 41.67
        & 20.39
        \\
        GT poses
        & 77.77
        & 34.16
        & \textbf{65.86}
        & 32.60
        \\

        \midrule

        \multicolumn{5}{@{}l}{\textit{(b) Predictor variants}} \\
        
        CAT-Seg~\cite{cho2024catseg}
        & 77.88
        & \textbf{38.20}
        & 65.64
        & 32.79
        \\
        NACLIP~\cite{hajimiri2025naclip}
        & 77.88
        & 29.32
        & 65.64
        & 31.82
        \\

        \bottomrule
    \end{tabular}
    \vskip-4ex
\end{table}

\section{Conclusion}
In this work, we presented PanOVOcc, a training-free framework for persistent open-vocabulary semantic occupancy mapping from native panoramic RGB-D sequences.
By integrating panoramic SLAM and open-vocabulary perception with long-term spatial voxel memory, PanOVOcc extends panoramic occupancy mapping from isolated local predictions to continuously updated, language-queryable global maps.
Experiments on the proposed benchmarks demonstrate improved geometric and semantic occupancy accuracy over the evaluated baselines, with particularly strong gains on major structural categories.
These findings highlight the value of combining omnidirectional sensing with persistent spatial memory for open-vocabulary embodied scene understanding.

\bibliographystyle{IEEEtran}
\bibliography{bib}

\end{document}